\documentclass[runningheads]{llncs}

\usepackage[T1]{fontenc}
\def\doi#1{\href{https://doi.org/\detokenize{#1}}{\url{https://doi.org/\detokenize{#1}}}}
\usepackage{graphicx}

\usepackage{listings}
\usepackage{comment}
\usepackage{booktabs}
\usepackage{amssymb}
\usepackage{pifont}
\usepackage{graphicx}
\usepackage[dvipsnames]{xcolor}
\usepackage{caption}
\usepackage{subcaption} 
\usepackage{xcolor}
\usepackage{multirow}
\usepackage{amsmath}
\usepackage{colortbl}
\usepackage{lipsum}
\usepackage{hhline}
\usepackage{kotex}
\usepackage[export]{adjustbox}
\usepackage{color,soul}
\usepackage{bm}
\usepackage[ruled,linesnumbered, vlined,algo2e]{algorithm2e} 
\usepackage{enumitem}
\usepackage[toc,page]{appendix}
\usepackage[normalem]{ulem}

\newcommand{\revision}{\textcolor{black}}
\newcommand{\xmark}{\ding{55}}%

\newcommand{\revisionf}[1]{\textcolor{black}{#1}}

\newcommand{\resthree}{$RN50_{[32]}$}
\newcommand{\ressix}{$RN50_{[64]}$}
\newcommand{\resonetwo}{$RN50_{[128]}$}
\newcommand{\restwotwo}{$RN50_{[224]}$}

\newcommand{\resonetwonine}{$RN50_{[129]}$}
\newcommand{\res}{$RN50$}
\newcommand{\MA}{MA }
\newcommand{\ID}{ID }
\newcommand{\MAs}{MAs }
\newcommand{\IDs}{IDs }
\newcommand{\MAdot}{MA}
\newcommand{\IDdot}{ID}
\newcommand{\IDsdot}{IDs}
\newcommand{\wrestwoeightone}{$WRN28$-$1_{[32]}$}
\newcommand{\wrestwoeightfive}{$WRN28$-$5_{[32]}$}
\newcommand{\wrestwoeightten}{$WRN28$-$10_{[32]}$}

\newcommand{\vggthree}{$VGG16_{[32]}$}
\newcommand{\vggsix}{$VGG16_{[64]}$}
\newcommand{\vggonetwo}{$VGG16_{[128]}$}
\newcommand{\vggtwotwo}{$VGG16_{[224]}$}

\newcommand{\vggninetwotwo}{$VGG19_{[224]}$}
\newcommand{\vggsixtwotwo}{$VGG16_{[224]}$}
\newcommand{\resoneoonetwotwo}{$RN101_{[224]}$}

\newcommand{\PP}[1]{
\vspace{2px}
\noindent{\bf \IfEndWith{#1}{.}{#1}{#1.}}
}

\newcommand{\squashlist}{
\begin{itemize}[noitemsep,nolistsep]
\setlength{\itemsep}{-0pt}
}
\newcommand{\squashend}{
  \end{itemize}
}

\makeatletter
\newcommand{\@chapapp}{\relax}%
\makeatother
\newcolumntype{C}[1]{>{\centering\arraybackslash}p{#1}}
\usepackage{makecell}

\usepackage[capitalize]{cleveref}

\begin{document}
\title{Precise Extraction of Deep Learning Models via Side-Channel Attacks on Edge/Endpoint Devices}
\titlerunning{Precise Extraction of Deep Learning Models}
%

\author{Younghan Lee\inst{1}\and
Sohee Jun\inst{1} \and
Yungi Cho\inst{1} \and
Woorim Han\inst{1} \and \\
Hyungon Moon\inst{2}\textsuperscript{*} \and
Yunheung Paek\inst{1}\textsuperscript{*} \\
}

\authorrunning{Y. Lee et al.}
%
\institute{Seoul National University, Seoul, Republic of Korea\\ 
\and UNIST, Ulsan, Republic of Korea \\ 
\email{\{201younghanlee, soheejun12, rimwoo98, ypaek\}@snu.ac.kr, ygcho@sor.snu.ac.kr, hyungon@unist.ac.kr}\\
{$^{*}$: Correspondence should be addressed to H. Moon and Y. Paek.}
}
\maketitle              
\begin{abstract}
With growing popularity, deep learning (DL) models are becoming larger-scale, and only the companies with vast training datasets and immense computing power can manage their business serving such large models. Most of those DL models are proprietary to the companies who thus strive to keep their private models safe from the model extraction attack (MEA), whose aim is to steal the model by training surrogate models. Nowadays, companies are inclined to offload the models from central servers to edge/endpoint devices. As revealed in the latest studies, adversaries exploit this opportunity as new attack vectors to launch side-channel attack (SCA) on the device running victim model and obtain various pieces of the model information, such as the model architecture (MA) and image dimension (ID). Our work provides a comprehensive understanding of such a relationship for the first time and would benefit future MEA studies in both offensive and defensive sides in that they may learn which pieces of information exposed by SCA are more important than the others. Our analysis additionally reveals that by grasping the victim model information from SCA, MEA can get highly effective and successful even without any prior knowledge of the model. Finally, to evince the practicality of our analysis results, we empirically apply SCA, and subsequently, carry out MEA under realistic threat assumptions. The results show up to 5.8 times better performance than when the adversary has no model information about the victim model.


\end{abstract}

\section{Introduction}
\label{sec:intro}

\revisionf{Deep learning (DL)} models empower many commercial applications and are potentially worth millions of dollars~\cite{lorica2018state,TopCompa99:online,ribeiro2015mlaas}. Until now, most model architectures and topology have been publicly available, but as models become larger-scale, the increased training cost and difficulty drive companies to prohibit the competitors from creating a copy and taking the market share. The cost of training a \revisionf{DL} model comes from both the computational resources and training datasets. Recent studies have shown that the \textit{model extraction attack} (MEA), aiming to train a \textit{surrogate model} of similar performance with much less training cost, is a real threat to such efforts of protecting valuable \revisionf{DL} models~\cite{tramer2016stealing,papernot2017practical,correia2018copycat,orekondy2019knockoff,barbalau2020black,pal2020activethief,yu2020cloudleak}.
Unfortunately, black-box MEAs require tremendous computational resources and time overhead~\cite{hu2020deepsniffer}. To mitigate the amount of labor and increase the chances of success, they usually make certain unrealistic assumptions that give them pre-knowledge about the victim model. For instance, a typical assumption is that the surrogate model has the same or more complex model architecture and the same image dimension as the victim~\cite{tramer2016stealing,orekondy2019knockoff,pal2020activethief}.

The growing demand for on-device ML services is fulfilled by offloading their models to edge/endpoint devices \cite{li2018deeprebirth}, which the adversary may access physically or via network connections, to improve response times and save bandwidth.
All in all, this recent development in ML computing opened up a new opportunity that the adversary may exploit as attack vectors to wage \textit{side-channel attacks} (SCAs) on the machine running the victim model. For example, when the victim model and the adversary's application run on the same device, the cache memory may be shared between them, which renders the model vulnerable to the cache SCA~\cite{yan2020cache}.
After gathering run-time information leaked via the device hardware, SCA can provide the adversary essential information about the victim that includes \textit{model architecture} (\MAdot) and \textit{image dimension} (\IDdot) of a \revisionf{DL} model. Such information is essentially identical to the assumed prior knowledge necessitated for boosting black-box MEAs. This means that with the aid of SCA, MEA can still build a surrogate model better resembling the target even without unrealistic initial assumptions on the victim side. However, there is a stumbling block to the full utilization of SCA for MEA; SCA does not come for free but requires a great deal of cost and effort to obtain sufficient model information accurately~\cite{yan2020cache,hu2020deepsniffer}. 
To improve their work, SCAs usually need to make strong assumptions on their target systems, which could often be unrealistic or broken just by simple obfuscation techniques~\cite{zhang2013duppel}. Therefore, a practical and efficient way to use SCA for MEA should be to extract only the essential information required to gain enough knowledge rather than prying into the target device to collect all sorts of information ignorantly. In order to pinpoint such essential information, we must understand how each piece of information affects the performance of MEA.
Unfortunately, to the best of our knowledge, no studies have examined extensively the effects of different pieces in the collectable information on MEA. 
There are some preliminary studies demonstrating that MEA is robust to the difference in certain model features, such as \MAdot, as long as the surrogate model's complexity is high enough~\cite{orekondy2019knockoff,pal2020activethief}.  Nonetheless, there has been no analytical report on the significance of other types of model information like \ID in influencing the effectiveness of MEA. 

Our work is the first to present an empirical analysis of the effects of SCA on MEA by evaluating the relationship between the performance of MEA and the model information supplied by SCA. We endeavor to empirically verify the relationship with various settings such as datasets, attack query budget, and attack strategy. 
We delve into linking a particular type of model information with the outcomes of MEAs by investigating the correlation. We believe that our analytical report will give a glimpse of what types of model information are of more value to boost the performance of MEA relatively. Thus, it will enhance the efficacy of SCAs in their assistance to MEA by letting them concentrate on such valuable ones. \revisionf{In addition, we demonstrate the practicality of utilizing ID obtained from SCA to boost MEA by carrying out the experiment under realistic assumptions. The results achieve up to 5.8x much higher accuracy and fidelity than the adversary without any prior knowledge.} Consequently, we argue that our work paves the way for future (offensive and defensive) research in \revisionf{DL} model privacy by providing organized knowledge of correlation between existing MEAs with the SCA-supplied knowledge about the victim model. 
The following summarizes our contributions:
\squashlist
\item We analyze the effect of model information exposed by SCA on MEA with different settings: datasets, query budget, and attack strategy.

\item \revisionf{We demonstrate how accurately ID of \revisionf{DL} models can be estimated by SCA and perform subsequent MEA with estimated ID under realistic threat assumptions to evince the practicality of MEA with SCA.}

\item We provide an informative insight into improving the defense against MEA allied with SCA by identifying what parts of model information are to be obfuscated from SCA for maximum defense with minimum effort.
\squashend

\section{Background} \label{sec:back}

While MEA and SCA ultimately share the same goal of extracting the victim model of high value, their interpretations of a successful attack are different. MEA aims to obtain a replica of its victim model by copying the functionality. 
In contrast, SCA intends to extract the structural or architectural model information, including dimensions of layers and their topology. 

\textbf{MEA} strives to create at low cost a surrogate of a high-performance victim \revisionf{DL} model trained with a dataset of both high quality and large quantity. The most popular method relies on only querying and collecting the inference results from the victim model. The surrogate model is trained with the data used to query the victim with the classification result as its label. As the cost of extracting the model increases with querying more samples, recent studies focus mainly on selecting more valuable samples to reduce the query budget. Ideally, MEA must be carried out with a pure black-box setting where no model information are initially exposed to the adversary. However, in reality, to reduce the amount of labor and increase the chances of their success, many MEA techniques are evaluated under certain assumptions that they already have prior knowledge (\ID and \MAdot) about the victim model. We find that such assumptions are often unrealistic in practice as some knowledge can not be available to adversaries in a black-box setting.

\textbf{SCA} has long been studied by cyber security researchers for many decades. In recent years, the use of SCA has been broadened to extract the architectural information of valuable \revisionf{DL} models.
Although SCA requires the assumption that adversaries gain access to hardware resources in the machine running the victim models, this assumption is deemed plausible these days, as discussed earlier. SCA attempts to collect model information by exploiting vulnerabilities in the underlying hardware. Previous studies have shown that SCA can collect mainly two types of model information, \ID and \MAdot. We note here that these are the same pieces of information somehow given to the adversary by assumptions. Further details of SCA is explained in~\cref{5_SCA-backed}.
\section{Related Work}
\label{sec:related}

\subsection{Model Extraction Attack}
\label{sec:MEA}
\textit{KnockoffNets}~\cite{orekondy2019knockoff} is one of the early MEA technique which is designed under black-box setting. The query samples are selected randomly from out-of-distribution attack dataset as the adversaries are unaware of the training dataset used by the victim model. Finally, the output label from the victim model is paired with the query data (i.e., \textit{re-labeled image}) and used as training dataset for the surrogate model which will exhibit a similar functionality as the victim.

\textit{ActiveThief}~\cite{pal2020activethief} proposed another method called \textit{uncertainty} which is based on the confidence vector of the query samples. The intuition behind this approach is that to extract the classification functionality of the victim model, it is beneficial to concentrate on the query samples that will lie near the decision boundary. By doing so, the surrogate can be trained much more quickly with fewer query samples to reach the victim's classification accuracy.

\textit{KnockoffNets} and \textit{ActiveThief} discuss briefly how the knowledge of \MA affects their performance, and conclude that MEAs are relatively robust to (or regardless of) the choice of MA for their surrogates if the MA complexity is high enough.
In other words, MEA can achieve good performance as long as the complexity of a surrogate is sufficiently high (usually, higher than the victim model). This implies that if SCA reveals the exact complexity of \MAdot, the adversary can set up the surrogate with MA of optimal complexity to maximize the effectiveness of MEA. Regarding \IDdot, KnockoffNets and ActiveThief are evaluated under a strong assumption that the adversary is aware of \ID of the victim and thus can set up the surrogate with the same \IDdot. They ignorantly assume that this essential pre-knowledge could be offered to attackers by convention, and none of them show how the attackers obtain such information. In this paper, we argue that this assumption can be fulfilled by employing SCA to decide the victim model \ID in a deterministic manner and will empirically prove that it is indeed indispensable to the success of MEA.

\subsection{Side-channel Attack} 
\textit{Cache Telepathy}~\cite{yan2020cache} utilized cache side-channel attack in estimating the architectures in MLaaS (Amazon SageMaker~\cite{amazon}, Google ML Engine~\cite{google}) platforms. By inferring the size of each layer's input matrix, they deduce the \ID of the model. Also, they can infer the \MA by identifying the topology of layers. To evaluate the accuracy of their method, the extracted structure is compared to the model's original structure. However, they do not show how useful their extracted model information is to boost MEA, which is necessary to understand the individual effect of each piece of information on black-box MEA.
\section{\revisionf{Analysis} of the Effects of Model Information}
\label{analysis}

\begin{figure}[!h]
\begin{minipage}[b]{\columnwidth}
  \centering\raisebox{\dimexpr \topskip-\height}{%
  \includegraphics[width=0.9\textwidth]{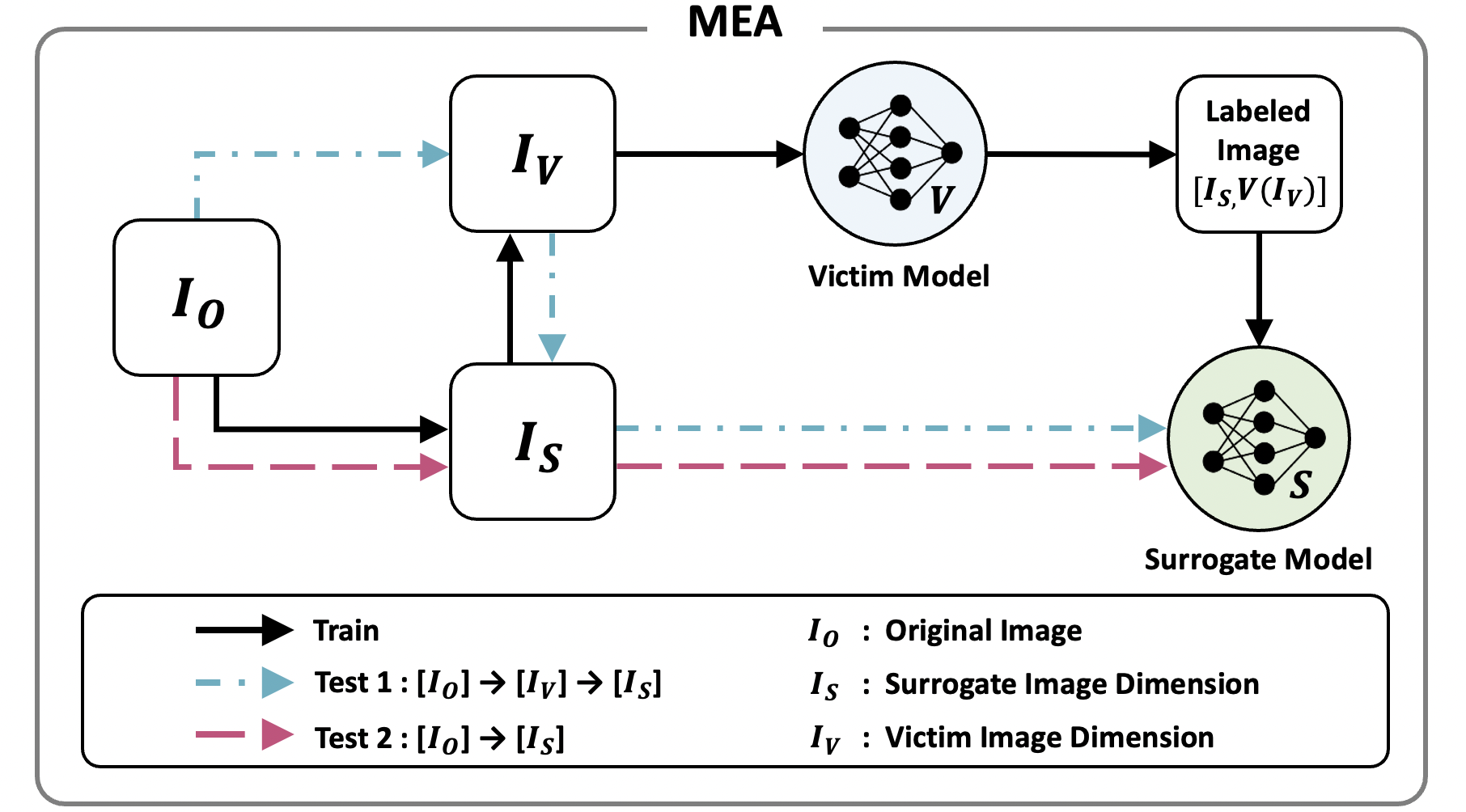}}
\caption{Flowchart of Training and Evaluating Surrogate Models \revision{through MEA}}
\label{fig:flowchart}
\end{minipage}\hfill
\end{figure}

As described just above, it is evident that an attacker can acquire by SCA the model information (i.e., \IDdot, \MAdot) pivotal to MEA which otherwise would have to rely on somewhat ungrounded assumptions to attain its goal.
In this work, we conduct a comprehensive \revisionf{analysis} to understand the effects of SCA on MEA. For this, we evaluate the performance results of MEA for various configurations of \ID and \MAdot, and find the following relationships. \textbf{R1:}~Effectiveness of MEA vs. model information (\ID and \MAdot) of the victim. \textbf{R2:}~\textbf{R1} vs. \revisionf{analysis} settings ($a$: datasets, $b$: attack query budget, $c$: attack strategy).

\subsection{Training \& Evaluation}
\revision{In this subsection, we elaborate~\cref{fig:flowchart} in further detail, in which the process of MEA is illustrated by training and evaluating surrogate models.}

\begin{figure}[htp]
\begin{subfigure}[b]{\columnwidth}
\makebox[1pt][r]{\makebox[27pt]{\raisebox{33pt}{\rotatebox[origin=c]{0}{{$\Biggl[$ \large{$$\bm{$I_s$}$$} \textbf{=}}}}}}%
\makebox[0pt][r]{\makebox[-55pt]{\raisebox{67pt}{\rotatebox[origin=c]{0}{\textbf{Image}}}}}%
\centering
\includegraphics[width=0.25\columnwidth]{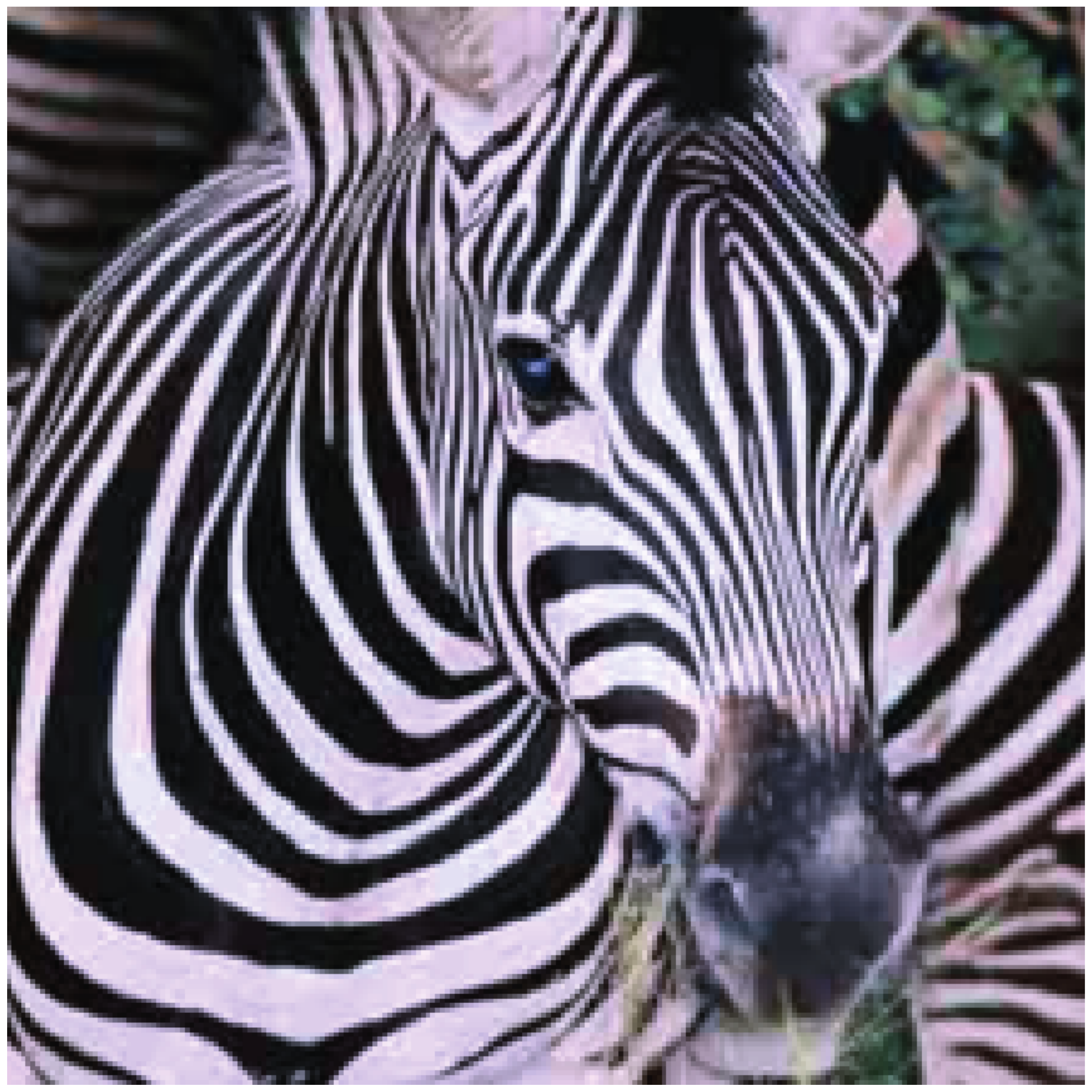}
\makebox[1pt][r]{\makebox[-3pt]{\raisebox{20pt}{\rotatebox[origin=c]{0}{\textbf{,}}}}}%
\hspace{4.5em}%
\makebox[0pt][r]{\makebox[-60pt]{\raisebox{67pt}{\rotatebox[origin=c]{0}{\textbf{Label}}}}}%
\makebox[1pt][r]{\makebox[27pt]{\raisebox{33pt}{\rotatebox[origin=c]{0}{\large{$$\bm{$V$}$$} $\biggl($\large{$$\bm{$I_v$}$$} \textbf{=}}}}}%
\centering
\includegraphics[width=0.25\columnwidth]{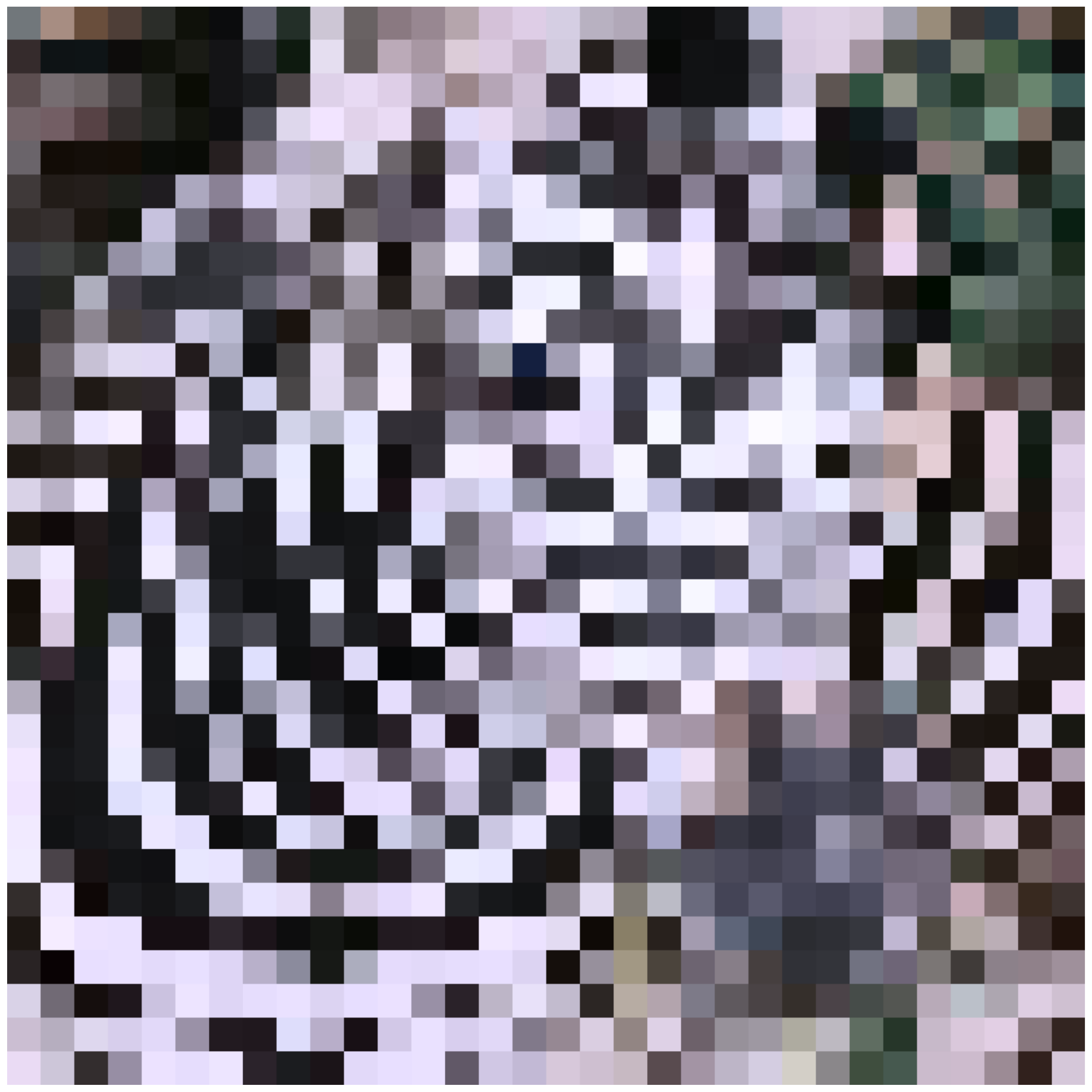}
\makebox[1pt][r]{\makebox[-8pt]{\raisebox{33pt}{\rotatebox[origin=c]{0}{$\biggr)$$\Biggr]$}}}}
\end{subfigure}%
\caption{Example of Re-labeled Image Example (Surrogate [224], Victim [32]) }
\label{fig:zibra}
\end{figure}

\textbf{Surrogate \revision{Model} Training.} 
\revision{By employing MEA,}~we trained the surrogate model following the flow depicted as the solid black line in~\cref{fig:flowchart}. We first resize the original image $I_O$ \revision{(\ID of original image)} to $I_S$ (\ID of the surrogate \revision{model}) and resize once more to $I_V$ (\ID of the victim \revision{model}). This is a realistic attack scenario as the adversary is unaware of the victim's image dimension
and is likely to query the victim with $I_S$. The confidence score that the victim returns is $V(I_V)$ and the surrogate model is trained with the \textit{re-labeled image} with the new label, [$I_S$, $V(I_V)$]. A discrepancy in $I_S$ and $I_V$ causes the \textit{label mismatch}, which is a difference in the classification score of $I_S$ and $I_V$ given by the victim model. ~\cref{fig:zibra} visually illustrates the \textit{re-labeled image} where the new label of $I_S$ is given by the classification score of the victim with lower \ID ($I_V$).

\textbf{Surrogate \revision{Model} Evaluation.} 
\revision{To measure the effectiveness of MEA,}~we evaluated the surrogate model with two separate tests as shown in~\cref{fig:flowchart}. Test 1 resizes $I_O$ to $I_V$ before eventually arriving at $I_S$, and test 2 resizes $I_O$ directly to $I_S$. \revision{Test 1 is similar to the current method of evaluation where model information including \ID(i.e., $I_V$) is known to adversaries.} Test 2 can be considered as a more realistic way of measuring the accuracy of the surrogate model as \ID of the victim is unknown to the adversary in reality. We note that when \ID of the surrogate is the same as that of the victim, there is no difference between test 1 and test 2 as $I_S = I_V$.
The result tables~(\cref{tbl:60000_perf_acc},~\cref{tbl:activethief}, \cref{tbl:arch_perf}, \cref{tbl:arch_perf2}, \cref{tbl:sub_mea}) include both tests, and the reported surrogate accuracy denotes the best accuracy among the ones measured every five epochs during the training.

\subsection{\revisionf{Analysis} settings}

\textbf{Image Dimension (\IDdot).} To understand the relationship between \IDs and the MEA effectiveness, we optimize the models with various \IDs to achieve the best accuracy for the victim models. We choose ResNet-50~\cite{he2016deep} (Additional result for VGG16~\cite{simonyan2014very} in~\cref{appendix:d}) as the architectures of both the victim and surrogate models. The \ID is represented as a subscript (i.e., \ressix ~represents ResNet-50 model optimized for 64x64x3 images).

\begin{table}[!h]
    \centering
    \caption{Dataset Configuration}
    \def\arraystretch{1}
\resizebox{\columnwidth}{!}{
\begin{tabular}{c|c||c|c|c|c|c}
\toprule
& 
 
\multirow{1}{*}{Dataset}& 
\multirow{1}{*}{Classes} & 
Train Samples&
Test Samples&
Original Image ($I_O$)&
\multirow{1}{*}{Analysis} \\

\midrule 
\hline
\multirow{4}{*}{Victim}&

Indoor\cite{indoor67} &  67 & 14,280 & 1,340  & 224x224x3& \IDdot \& \MA\\

 & Caltech-256\cite{caltech256} & 256 & 23,380 & 6,400 & 224x224x3 & \IDdot \& \MA\\ 
 & CUB-200\cite{cubs200} &  200 & 5,994 & 5,794 & 224x224x3 & \IDdot \\ 
 & CIFAR-100\cite{cifar10} &  100 & 50,000 & 10,000  & 32x32x3& \MAdot\\
\hline

\multirow{2}{*}{Attack} & ImageNet\cite{ILSVRC15}  &  1,000 & 1.2M & 150,000 & 224x224x3 &
\IDdot \& \MA \\
 &  OpenImages\cite{kuznetsova2020open} &  600 & 1.74M & 125,436 & 224x224x3 & \IDdot \\

\bottomrule
\end{tabular}

}

    \label{tbl:datasets_info}
\end{table}

\textbf{Datasets.} For \ID \revisionf{analysis}, the victim models are trained with three widely used datasets, as shown in~\cref{tbl:datasets_info}. To achieve a realistic and high-accuracy model, they are trained by a transfer learning method with a pre-trained model by ISLVRC-2012 (ImageNet)~dataset. The accuracy of the trained victim models is shown in the second column of~\cref{tbl:60000_perf_acc}. For the attack query dataset, we follow the common assumption that an adversary is unaware of the dataset used for training the victim model. The \revisionf{analysis} is performed with \textit{out-of-distribution} datasets, ImageNet and OpenImages, with enough samples for a large query budget \revisionf{analysis}. We note that OpenImages is an unbalanced dataset where there are uneven number of samples for each image class. For \MA analysis, along with three forementioned datasets, we added CIFAR-100 as shown in~\cref{tbl:datasets_info} and only ImageNet is used as an attack query dataset. The accuracy of the victim models is shown in the second column of~\cref{tbl:arch_perf} and~\cref{tbl:arch_perf2}.

\textbf{Attack Query Budget.}
Various attack query budgets are used for \ID \revisionf{analysis} (i.e., 30k, 60k, and 90k). It is designed to verify if the relationship between \ID of the victim and the effectiveness of MEA changes over different query budgets.

\textbf{Attack Strategy.}
For \ID \revisionf{analysis}, two attack strategies are implemented. We use a random sampling strategy based on KnockoffNets and \textit{uncertainty} method from ActiveThief because they exhibit arguably the best performance in MEA. For \MA analysis, KnockoffNets was implemented.

\textbf{Model Architecture (\MAdot).}
 to understand the relationship between \MA and the effectiveness of MEA, various \MAs of different complexities are used to train the victim and surrogate models: WideResNet-28-k~\cite{zagoruyko2016wide} with different k values, VGG16, VGG19, ResNet-50 and ResNet-101. (Details in~\cref{appendix:b}).

\subsection{Effect of Image Dimension (\IDdot)}
\label{effect_of_ID}

{\textbf{Model Extraction Attack Result.}}
\cref{tbl:60000_perf_acc} depicts the result of MEA with the relative accuracy (i.e., accuracy of surrogate model relative to that of the victim which is 1x) as the effectiveness metric. The grey colored boxes denote that \IDs of the victim and the surrogate are the same. The bold type represents the best accuracy among surrogate models of different \IDsdot. The result confirms that matching the \ID of the victim and surrogate is vital to maximize the efficacy of MEA. Among the total of 48 cases in the grey colored boxes per metric, 92\% (44/48) achieve the best accuracy. We find only few cases in the upper right triangular matrix where the surrogate with a higher \ID achieved the same or better performance which is at most 3\% higher. However, when \ID is different, the accuracy of surrogate trained with higher \ID ($I_S > I_V$) drop by 0.72x at worst case (average drop of 0.24x). The decrease is more significant when the surrogate is trained with smaller \ID ($I_S < I_V$), showing 0.73x in accuracy at worst case (average drop of 0.43x). Between the effectiveness measured by test 1 and test 2 (excluding the cases where $I_s = I_V$), the result from test 1 is higher for 71\% ($51/72$) of the total cases. This instance can be explained by \textit{label mismatch} which is caused by the fact that the new label of $I_S$ is given by $I_V$. When the relative accuracy is measured by test 1, the influence of \textit{label mismatch} is diminished as $I_S$ is transformed from $I_V$ unlike test 2 in which $I_S$ is transformed from $I_O$ directly. In short, the effectiveness of MEA is maximized when the surrogate model's \ID matches the victim model's. (Ablation study results in~\cref{appendix:d})

\textbf{Datasets.} We examine if the trend described above is consistent throughout various datasets. Three different datasets are used for victim models and two attack query datasets for training the surrogate model.~\cref{tbl:60000_perf_acc} shows MEA is most effective when the surrogate's \ID is identical for all three victim model datasets which have different number of classes. The effectiveness is generally higher with ImageNet which achieve higher relative accuracy in 68\% ($65/96$) of total cases with the average rise of 1.4\%. This phenomenon is due to the fact that the victim models are pre-trained with ImageNet dataset. However, it is important to note that the trend continues for both attack query datasets (i.e., Imagenet and OpenImages).

\begin{table*}[h]
\centering
\caption{\ID \revisionf{Analysis (Datasets).} Effectiveness (Relative Accuracy) of MEA (KnockoffNets) with query-budget-60k}

\def\arraystretch{1.2}
\resizebox{\linewidth}{!}{
    \begin{tabular}{c|c|c||c|cc|cc|cc|cc}
    \toprule
     \multicolumn{3}{c||}{{Victim Model}} &
     \multicolumn{9}{c}{{Surrogate Model}}\\
     \midrule
      
          \multirow{2}{*}{Dataset} & a
          \multirow{2}{*}{Accuracy} & 
          \multirow{2}{*}{Model} & 
           \multirow{2}{*}{Attack Query} &
          \multicolumn{2}{c|}{\resthree} & 
          \multicolumn{2}{c|}{\ressix} & 
          \multicolumn{2}{c|}{\resonetwo} & 
          \multicolumn{2}{c}{\restwotwo}  \\
    \cline{5-12}

    &&&&
    Test 1&
    Test 2&
    Test 1&
    Test 2&
    Test 1&
    Test 2&
    Test 1&
    Test 2\\

    \midrule
    \hline

     %
    \multirow{2}{*}{Indoor67} & 
    \multirow{2}{*}{64.78\% (1x)} &
    \multirow{6}{*}{\resthree} & 
    ImageNet &    
   
    {\textbf{0.88x }}\cellcolor{gray!15} & {0.88x }\cellcolor{gray!15} &

    0.63x & \textbf{0.91x}  &

    0.59x  & 0.50x &

   0.43x  & 0.16x\\

    &&&
    OpenImages &    
   
    {\textbf{0.91x}}\cellcolor{gray!15} & {\textbf{0.91x}}\cellcolor{gray!15} &
    
     0.69x & \textbf{0.91x} &

    0.62x & 0.44x &
    
   0.46x & 0.17x\\
   
    \cline{1-2}
     \cline{4-12}  
    \hhline{~|~|~|~|-}

    \multirow{2}{*}{Caltech-256} & 
    \multirow{2}{*}{66.56\% (1x)} &
    &   
    ImageNet &   
     
    {\textbf{0.96x}}\cellcolor{gray!15} & 0.96x\cellcolor{gray!15} &

    0.78x & \textbf{0.97x} &
    
    0.75x &  0.61x &
    
    0.59x &  0.28x\\
    &
    &
    &   
    OpenImages & 
    {\textbf{0.94x}}\cellcolor{gray!15} & 0.94x\cellcolor{gray!15} &

    0.75x & \textbf{0.95x} &
    
    0.66x& 0.53x &
    
    0.47x &  0.23x \\      
    \cline{1-2}
     \cline{4-12}  
     \hhline{~|~|~|~|-}

    \multirow{2}{*}{CUB-200} & 
    \multirow{2}{*}{67.02\% (1x)} &
    &
    ImageNet &   
   
    {\textbf{0.86x}}\cellcolor{gray!15} & {\textbf{0.86x}}\cellcolor{gray!15} &

     0.62x & 0.80x & 
     
    0.51x & 0.40x &
    
    0.35x & 0.15x\\
   
    &&&
    OpenImages & 
    
    {\textbf{0.83x}}\cellcolor{gray!15} & {\textbf{0.83x}}\cellcolor{gray!15} &

    0.56x & 0.73x & 
    
    0.48x & 0.35x &
    
    0.31x & 0.14x\\
    \midrule
    
    
    %
    \multirow{2}{*}{Indoor67} & 
    \multirow{2}{*}{72.99\% (1x)} &
    \multirow{6}{*}{\ressix} & 
    ImageNet &   
    0.33x & 0.28x &
    
    {\textbf{0.94x}}\cellcolor{gray!15} & {\textbf{0.94x}}\cellcolor{gray!15} &
    
    0.77x & 0.87x &

    0.69x & 0.49x\\
    
    &&&
    OpenImages & 
    0.35x & 0.29x &
    
    {\textbf{0.96x}}\cellcolor{gray!15} & {\textbf{0.96x}}\cellcolor{gray!15} &

    0.85x & 0.91x &

    0.71x & 0.53x \\
    
    \cline{1-2}
     \cline{4-12}  
     \hhline{~|~|~|~|~|-}

    \multirow{2}{*}{Caltech-256} & 
    \multirow{2}{*}{76.81\% (1x)} &
    &   
    ImageNet &   
    0.51x & 0.48x & 
    
    {\textbf{0.99x}}\cellcolor{gray!15} & {\textbf{0.99x}}\cellcolor{gray!15} & 

    0.90x &  0.96x &
    
    0.85x &  0.72x \\

    &&&   
   OpenImages & 
    0.48x & 0.45x & 
    
    {\textbf{0.97x}}\cellcolor{gray!15} &{\textbf{0.97x}}\cellcolor{gray!15} & 
    
    0.87x & 0.94x &
    
    0.78x &  0.69x \\  
    
    \cline{1-2}
     \cline{4-12}  
    \hhline{~|~|~|~|~|-}

    \multirow{2}{*}{CUB-200} & 
    \multirow{2}{*}{77.89\% (1x)} &
    &
    ImageNet &   
    0.15x & 0.13x & 
    
    {\textbf{0.88x}}\cellcolor{gray!15} & {\textbf{0.88x}}\cellcolor{gray!15} &
    
    0.66x & 0.79x &
    
    0.58x & 0.40x\\
   
    &&&
   OpenImages & 
    0.13x & 0.11x & 
    
    {\textbf{0.82x}}\cellcolor{gray!15} & {\textbf{0.82x}}\cellcolor{gray!15} &
    
    0.65x & 0.76x &
    
    0.55x & 0.37x \\
    
    \midrule
    

    %
    \multirow{2}{*}{Indoor67} & 
    \multirow{2}{*}{67.24\% (1x)} &
    \multirow{6}{*}{\resonetwo} & 
     ImageNet &  
    0.33x & 0.22x & 
     
    0.82x & 0.78x & 
     
    {\textbf{0.97x}}\cellcolor{gray!15} & {\textbf{0.97x}}\cellcolor{gray!15} & 
    
    0.95x & 0.94x\\
    
    &&&
    OpenImages & 
    0.33x & 0.22x & 
     
    0.84x & 0.80x & 
     
    {\textbf{1.00x}}\cellcolor{gray!15} & {\textbf{1.00x}}\cellcolor{gray!15} &

    0.96x & 0.96x \\
    \cline{1-2}
     \cline{4-12}  
    \hhline{~|~|~|~|~|~|-}

    \multirow{2}{*}{Caltech-256} &  
    \multirow{2}{*}{76.75\% (1x)} &&
    ImageNet &   
    0.44x & 0.43x &  
    
    0.78x & 0.75x &
    
   {\textbf{0.99x}}\cellcolor{gray!15} & 
   {\textbf{0.99x}}\cellcolor{gray!15} &
  
    0.97x & 0.97x\\
    
    &&&
   OpenImages & 
    0.43x & 0.42x & 

    0.76x & 0.73x &

    {\textbf{0.97x}}\cellcolor{gray!15} & 0.97x\cellcolor{gray!15} &
  
    0.95x & \textbf{0.98x}\\
    
    \cline{1-2}
     \cline{4-12}  
    \hhline{~|~|~|~|~|~|-}

    \multirow{2}{*}{CUB-200} & 
    \multirow{2}{*}{77.44\% (1x)} &&  
    ImageNet &  
    0.21x & 0.15x &

    0.64x & 0.59x &

   {\textbf{0.91x}}\cellcolor{gray!15} & {\textbf{0.91x}}\cellcolor{gray!15} &

    0.86x & 0.87x \\
    
    &&&  
    OpenImages & 
    0.18x & 0.13x & 
    
    0.60x & 0.56x &
    
    {\textbf{0.88x}}\cellcolor{gray!15} & {\textbf{0.88x}}\cellcolor{gray!15} &
    
    0.83x & 0.84x\\
    \midrule

    \multirow{2}{*}{Indoor67} & 
    \multirow{2}{*}{73.51\% (1x)} &
    \multirow{6}{*}{\restwotwo} &  
    ImageNet &  
    0.26x &0.25x &
  
    0.66x & 0.67x &
 
    0.90x & 0.87x &
  
    {\textbf{0.92x}}\cellcolor{gray!15} & {\textbf{0.92x}}\cellcolor{gray!15}\\
    
    &&&  
    OpenImages & 
    0.26x & 0.23x & 
     
    0.69x & 0.69x &
     
    0.92x & 0.90x &
     
    {\textbf{0.97x}}\cellcolor{gray!15} & {\textbf{0.97x}}\cellcolor{gray!15} \\
    \cline{1-2}
     \cline{4-12}  
    \hhline{~|~|~|~|~|~|~|-}

    \multirow{2}{*}{Caltech-256} &  
    \multirow{2}{*}{78.11\% (1x)} &&  
    ImageNet &  
   0.36x & 0.39x &
    
    0.78x & 0.75x &
     
    0.95x & 0.92x &
     
    {\textbf{1.00x}}\cellcolor{gray!15} & {\textbf{1.00x}}\cellcolor{gray!15}  \\

    &&&  
  OpenImages & 
    0.34x & 0.38x & 
     
    0.74x & 0.73x &
     
    0.92x & 0.90x &
     
    {\textbf{0.99x}}\cellcolor{gray!15} & 
    {\textbf{0.99x}}\cellcolor{gray!15}  \\
    \cline{1-2}
       \cline{4-12}  
    \hhline{~|~|~|~|~|~|~|-}

    \multirow{2}{*}{CUB-200} & 
    \multirow{2}{*}{78.17\% (1x)} &&
     ImageNet &   
    0.17x & 0.16x &
     
    0.53x & 0.52x &
     
    0.78x & 0.78x &
     
    {\textbf{0.89x}}\cellcolor{gray!15} & {\textbf{0.89x}}\cellcolor{gray!15} \\
    &&&  
    OpenImages & 
    0.15x & 0.14x & 
    
    0.48x & 0.45x &
     
    0.74x & 0.71x &
     
    {\textbf{0.85x}}\cellcolor{gray!15} & 
    {\textbf{0.85x}}\cellcolor{gray!15} \\


    \bottomrule
    \end{tabular}%

}
  
\label{tbl:60000_perf_acc}
\end{table*}

\begin{figure}[!ht]
    \begin{subfigure}[b]{\columnwidth}
        \centering
        \begin{subfigure}[t]{0.8\columnwidth}
        \includegraphics[width=\columnwidth]{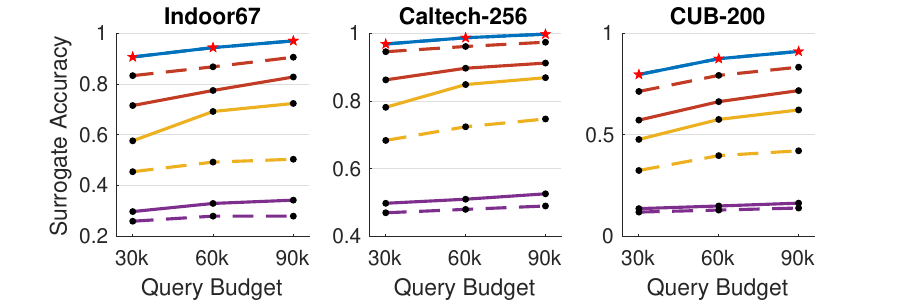}
        \end{subfigure} 

        \subcaption{Victim [64]}
    \end{subfigure}  
    
    \begin{subfigure}[b]{\columnwidth}
        \centering
        \begin{subfigure}[t]{0.8\columnwidth}
        \includegraphics[width=\columnwidth]{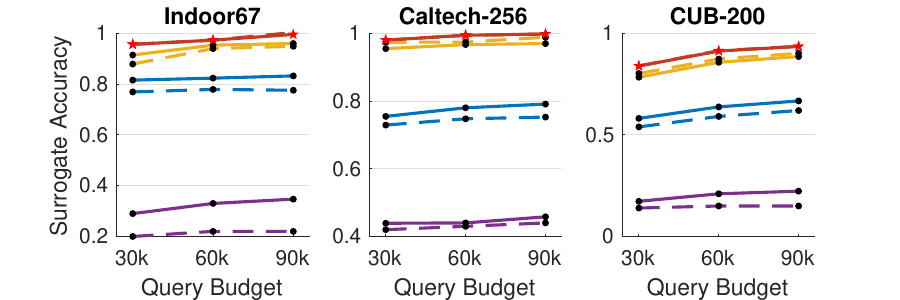}
        \end{subfigure} 

        \subcaption{Victim [128]}
    \end{subfigure}  

    \begin{subfigure}[b]{\columnwidth}
        \centering
        \begin{subfigure}[t]{0.8\columnwidth}
        \includegraphics[width=\columnwidth]{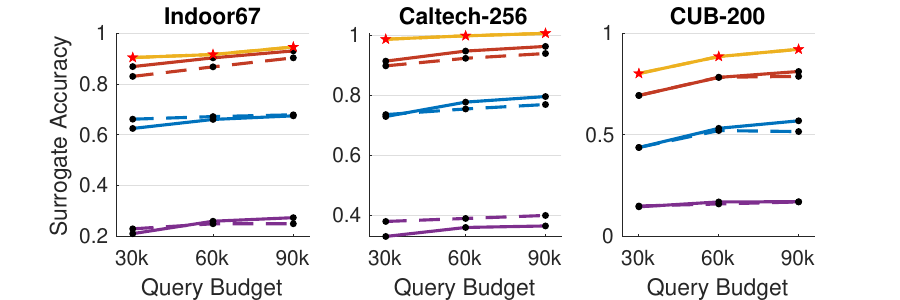}
          \end{subfigure}

        \subcaption{Victim [224]}
    \end{subfigure} 
    \hfill
    \begin{subfigure}[t]{1\columnwidth}
        \centering
        \includegraphics[width=0.8\columnwidth]{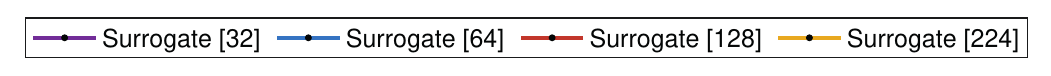}
    \end{subfigure} 
\caption{ID Analysis (Attack Query Budget). Effectiveness (Relative Accuracy) of MEA (KnockoffNets with ImageNet) for test 1 (solid line) \& test 2 (dotted line)}
\label{fig:TC12_BEST_BALANCE}
\end{figure}

\textbf{Attack Query Budget.} \revisionf{\cref{fig:TC12_BEST_BALANCE} illustrates 
changes in the effectiveness of MEA over various query budgets (30k, 60k, and 90k)} The surrogate model that matches the victim's \ID is marked with a red star marker at each query budget. The result shows that matching the surrogate's \ID to the victim model is always beneficial through various query budgets. Also, we note that even with a much less query budget, a higher accuracy can be attained when \ID is matched. In some cases, query-budget-30k with the same \ID can achieve a better accuracy than query-budget-90k with a different \IDdot. Moreover, as the cost of MEA increases as the query budget increases, the adversary can save a huge amount of cost just by training the surrogate with the same \ID as the victim.

\textbf{Attack Strategy.} We implement ActiveThief to verify if using a different attack strategy consorts with the phenomena observed in the previous \revisionf{analysis}. The attack is carried out with 2k initial seed samples and by sampling new 2k samples for 9 additional rounds (i.e., query-budget-20k).~\cref{tbl:activethief} shows the similar result that the surrogates trained with the identical \ID achieve the best accuracy.

\begin{table}[h]
\centering
    \caption{\ID \revisionf{Analysis} (Attack Strategy). The Effectiveness (Relative Accuracy) of MEA (ActiveThief with ImageNet) with query-budget-20k}
    \def\arraystretch{1.2}
\resizebox{\columnwidth}{!}{
    \begin{tabular}{c|c|c||cc|cc|cc|cc}
    \toprule
     \multicolumn{3}{c||}{{Victim Model}} &
     \multicolumn{8}{c}{{Surrogate Model}}\\
     \midrule
    
          \multirow{2}{*}{Dataset}&
           \multirow{2}{*}{Accuracy}&
          \multirow{2}{*}{Model}&

          \multicolumn{2}{c|}{\resthree} &  
          \multicolumn{2}{c|}{\ressix} & 
          \multicolumn{2}{c|}{\resonetwo} & 
          \multicolumn{2}{c}{\restwotwo}  \\
          
         \cline{4-11}
          &&&
          Test 1&
          Test 2&
          Test 1&
          Test 2&
          Test 1&
          Test 2&
          Test 1&
          Test 2\\
          
    \midrule
    \hline

    \multirow{4}{*}{Indoor67} & 
   \multirow{1}{*}{64.78\% (1x)}&
    \multirow{1}{*}{\resthree} &    
    {\textbf{0.82x}}\cellcolor{gray!15} & 
    {\textbf{0.82x}}\cellcolor{gray!15} &
    0.34x & 0.30x &
    0.48x & 0.44x &
    0.46x & 0.16x \\
    \cline{2-11}
    \hhline{~|~|~|~|-|-}

    &
        \multirow{1}{*}{72.99\% (1x)} &
    \multirow{1}{*}{\ressix} &    

    0.31x & 0.27x &
    {\textbf{0.90x}}\cellcolor{gray!15} & 
    {\textbf{0.90x}}\cellcolor{gray!15} &
    0.70x & 0.86x &
    0.65x & 0.50x \\
    \cline{2-11}
    \hhline{~|~|~|~|~|~|-|-}
    
    &
        \multirow{1}{*}{67.24\% (1x)} &
    \multirow{1}{*}{\resonetwo} &

     0.28x & 0.21x &
    0.78x & 0.75x &
    {\textbf{0.95x}}\cellcolor{gray!15} & {\textbf{0.95x}}\cellcolor{gray!15} &
    0.90x & 0.91x\\
    \cline{2-11}
    \hhline{~|~|~|~|~|~|~|~|-|-}

    &
     \multirow{1}{*}{73.51\% (1x)} &
    \multirow{ 1}{*}{\restwotwo} &  
    
     0.17x & 0.23x &
    0.60x & 0.65x &
    0.85x & 0.84x &
    {\textbf{0.88x}}\cellcolor{gray!15} & {\textbf{0.88x}}\cellcolor{gray!15} \\
    
    \bottomrule
 
    \end{tabular}%

}  
    \label{tbl:activethief}
\end{table}

\subsection{Effect of Model Architecture (\MAdot)}
\textbf{Model Extraction Attack Result.} In order to investigate how MA information of the victim model affects the effectiveness of MEA, we design the \revisionf{analysis} with various MA of different complexities. We note that in order to eliminate any effects of \IDdot, all models are set to have the same \IDdot. Therefore, the results shown in~\cref{tbl:arch_perf} and~\cref{tbl:arch_perf2} are from both test 1 and 2. Unlike ID \revisionf{analysis}, ~\cref{tbl:arch_perf} shows that MA of higher complexity achieves better accuracy than MA of the same complexity. Also,~\cref{tbl:arch_perf2} reveals a similar results as previous studies mentioned in~\cref{sec:MEA}. While the adversary can benefit from knowing the same model architecture, the effect is relatively less significant for most cases. Attacking with \resoneoonetwotwo achieves equally high or higher relative accuracy compared to attacking with the same model architecture. We conclude that the effect of \MA on MEA becomes insignificant as long as the surrogate model's MA occupies a high complexity.

\begin{table}[!h]
\centering
    \caption{\MA \revisionf{Analysis} 1. The Effectiveness (Relative Accuracy) of MEA (KnockoffNets with ImageNet) with query-budget-20k}
    \def\arraystretch{1.1}
\resizebox{1\columnwidth}{!}{
    \begin{tabular}{c|c|c||c|c|c}
    \toprule
     \multicolumn{3}{c||}{{Victim Model}} &
     \multicolumn{3}{c}{{Surrogate Model}}\\
     \midrule
    
    \multirow{1}{*}{Dataset}&
    \multirow{1}{*}{Accuracy}&
    \multirow{1}{*}{Model}&

         \multicolumn{1}{c|}{\wrestwoeightone} &  
        \multicolumn{1}{c|}{\wrestwoeightfive} &
        \multicolumn{1}{c}{\wrestwoeightten}
         \\

    \midrule
    \hline
    
    \multirow{3}{*}{CIFAR-100} & 
    \multirow{1}{*}{68.36\%  (1x)}& 
    \multirow{1}{*}{\wrestwoeightone} &    
    
    0.43x\cellcolor{gray!15}&

    0.56x& 
   
    {\textbf{0.57x}}\\

   \cline{2-6}
     \hhline{~|~|~|~|-|-}

     & 
    \multirow{1}{*}{77.95\% (1x)}& 
    \multirow{1}{*}{\wrestwoeightfive} &

    0.26x& 
  
    0.36x\cellcolor{gray!15}& 
 
    {\textbf{0.39x}}\\

   \cline{2-6}
    \hhline{~|~|~|~|~|-}

     & 
    \multirow{1}{*}{79.44\% (1x)}& 
    \multirow{1}{*}{\wrestwoeightten} &  
     
    0.26x& 
    
    0.37x& 
  
    {\textbf{0.39x}}\cellcolor{gray!15}\\

    \bottomrule
    \end{tabular}%
}

    \label{tbl:arch_perf}
\end{table}

\begin{table}[!h]
\centering
    \caption{\MA \revisionf{Analysis} 2. The Effectiveness (Relative Accuracy) of MEA (KnockoffNets with ImageNet) with query-budget-20k}
    \def\arraystretch{1.1}
\resizebox{1\columnwidth}{!}{
    \begin{tabular}{c|c|c||c|c|c|c|c}
    \toprule
     \multicolumn{3}{c||}{{Victim Model}} &
     \multicolumn{4}{c}{{Surrogate Model}}\\
     \midrule
    
    \multirow{1}{*}{Dataset}&
    \multirow{1}{*}{Accuracy}&
    \multirow{1}{*}{Model}&

    \multicolumn{1}{c|}{\vggsixtwotwo}&
    \multicolumn{1}{c|}{\vggninetwotwo} &
    \multicolumn{1}{c|}{\restwotwo}&
    \multicolumn{1}{c}{\resoneoonetwotwo}\\
    \cline{4-7}

    \hline
    \hline

    Indoor67&
    78.20\% (1x)&
    \multirow{2}{*}{\vggsixtwotwo}&
     \textbf{0.88x}\cellcolor{gray!15}&
    0.86x&	
    0.86x&
     \textbf{0.88x} \\
    \cline{1-2}
    \cline{4-7}
    \hhline{~|~|~|-|~|~|~}
    
    Caltech-256&
    83.06\% (1x)&
    &
    \textbf{0.94x}\cellcolor{gray!15}&
    0.92x&
    0.93x&
    \textbf{0.94x}\\

    \midrule

    Indoor67&
    78.13\% (1x)&
    \multirow{2}{*}{\vggninetwotwo}&
    0.83x&			
    \textbf{0.90x}\cellcolor{gray!15}&
    0.87x&
    \textbf{0.90x} \\
    \cline{1-2}
    \cline{4-7}
    \hhline{~|~|~|~|-|~|~}
    
    Caltech-256&
    85.77\% (1x)&
    &
    0.86x&
    0.93x\cellcolor{gray!15}&
    0.92x&
   \textbf{0.94x} & \\

    \bottomrule
    \end{tabular}%
}  

    \label{tbl:arch_perf2}
\end{table}
\section{Experiments}
\label{5_SCA-backed}
Our analysis, as explained in the previous section~(\cref{analysis}), suggests that an adversary knowing model information (i.e., ID) can boost the effectiveness of MEA. To estimate such information, an adversary can exploit SCA on local devices on which the victim model is running. 
\revisionf{In this section, we demonstrate end-to-end MEA with SCA \textit{without any prior knowledge} and confirm that our attack mechanism is realistic and highly effective to achieve  virtually the same ideal performance of MEA exhibited by the previous work assuming that they somehow manage to obtain model information before actual attacks.}

\subsection{Experimental Setups for MEA with SCA}
\label{experimentalsetup}
\begin{figure}[htp]
\begin{minipage}[b]{1\columnwidth}
  \centering\raisebox{\dimexpr \topskip-\height}{%
  \includegraphics[width=\textwidth]{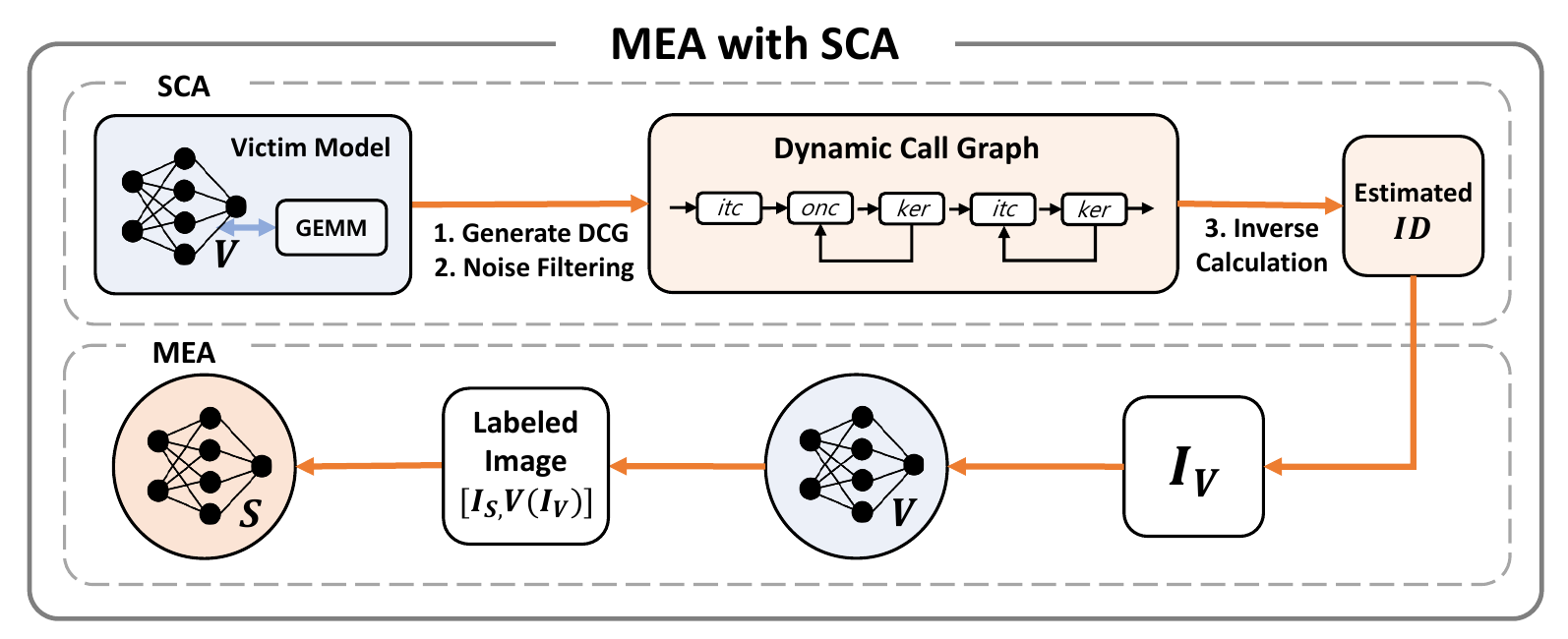}}
\caption{\revisionf{Flowchart of MEA with SCA. 1) DCG Generation, 2) Noise Filtering for DCG, 3) Inverse Calculation to estimate ID, 4) MEA with estimated ID}}
\label{fig:sca_mea}
\end{minipage}\hfill
\end{figure}

\textbf{Overview} Our SCA infers \ID from the
Generalized Matrix Multiply (GEMM), a commonly used building block of DL model implementation, operating in repeated loops for the matrix multiplication. If our SCA infers the number of iterations executed in each layer, it can compute the size of the layer's input matrix by multiplying the number with known constants. The target of our SCA, \IDdot, can be inferred in this way because \ID is equivalent to the first layer's input matrix size. Our implementation of SCA is based on \textit{Cache Telepathy}~\cite{yan2020cache} with modifications tailored for our purposes. \cref{fig:sca_mea} illustrates the process of \revisionf{MEA with SCA}. Firstly, we generate a dynamic call graph (DCG) that reflects the execution flow is generated by monitoring GEMM library with noise filtering mechanism. From the DCG, we can infer the number of iterations of loops executed in each layer. Secondly, ID of \revisionf{DL} model is estimated through the inverse calculation from the properties of the loops. Finally, estimated ID is used for subsequent MEA. Each step is described in more detail below.

\textbf{\revisionf{Threat Assumptions}} We performed SCA and MEA assuming that \revisionf{DL} model is a black-box and running on an edge/endpoint device. The adversary is not given direct access to the victim model, but only the prediction result is available. For cache-timing attack as a part of SCA, the tracing process by the adversary is running on the same processor as the victim model's process to capture the addresses \revision{which are managed by a process running in the background. Such scenario is realistic as \revision{the} DL model is off-loaded to the local device.} Also, the adversary is capable of analyzing linear algebra library used in \revision{the} \revisionf{DL} model such as OpenBLAS~\cite{zh19openblas} which is open-source. We assume that stride and padding are known to the adversary. For SCA experiment, we used OpenBLAS for GEMM library with a Linux machine running on a i7-9700 processor that has 8 cores, 64KB of L1 cache, 256KB of L2 cache, and 12MB of shared last-level cache. The system runs with 64GB of main memory.

\textbf{Dynamic Call Graph Generation} 
For the purpose of inferring the number of loop iterations, we used cache-timing attack as a part of SCA to infer the DCG that is composed of calls to three key functions, \texttt{itcopy}, \texttt{oncopy} and \texttt{kernel}. We chose these as key functions because each loop of our interest for generating DCG is composed of a specific sequence of such functions
as shown below.

\begin{itemize}[noitemsep,nolistsep,left=0pt..1em]
\setlength{\itemsep}{-0pt}
\item[] $L1$: \texttt{itcopy-oncopy-kernel-itcopy-kernel}
\item[] $L2$: \texttt{oncopy-kernel}
\item[] $L3$: \texttt{itcopy-kernel}
\end{itemize}

While the victim model is running, the adversary monitors the calls to these three key functions using \textit{Flush+Reload}, a common technique used for such purpose. By constantly monitoring the addresses of the key functions, we determine if one of the addresses has recently been accessed by measuring the access delay (i.e., cache hit or miss). Three properties of each loop can be deduced from DCG for the following procedure of MEA with SCA. 1) the number of iterations of loops (N), 2) short execution time (ST), 3) average execution time (AT). Short execution time is measured by the last function call and the average execution time is measured with all other calls. Further detail about the algorithm of DCG generation is described in~\cref{appendix:e}.

\textbf{Noise Filtering Mechanism}
We found that the DCG generation suffers from the inherent noise that SCA is prone to. We devised a noise filtering mechanism that is tailored for inferring a precise enough DCG exploiting the characteristics of GEMM computation in the target \revisionf{DL} architecture. While running, the execution time of each loop iteration is similar to each other except for the last one because each iteration processes inputs of similar size. For this reason, the interval between the function calls that we are monitoring is supposed to be similar to each other. Based on this, we filtered out duplicate observations of one function calls in two ways. First, we filtered out the function calls observed shortly after ($<$ 10 time intervals) the previous one, considering that the two adjacent observation is from a single function call. Second, we used the average interval between the function calls as a threshold and considered any calls to \texttt{itcopy} within the threshold as noise.

\textbf{Inverse Calculation Algorithm} In GEMM matrix multiplication (i.e., $m$ by $k$ and $k$ by $n$), $m$, $k$, $n$ represent input matrix, weight matrix, and output depth respectively and they are divided into loops by constant Q, P, and UNROLL which are pre-defined by the GEMM library. Therefore, after analyzing DCG to deduce three properties of each loop described above, we estimate ID of DL models by the inverse calculation. Firstly, $m$ value was calculated from properties of $L2$. As $m$ is divided by P to form $L2$ except for the last two iterations, which is operated by an unit of $(P+ m \mod P)/2$, $m$ can be obtained by the multiplication of P with the number of iterations of $L2$. Secondly, $n$ is divided by $3 \cdot \text{UNROLL}$ to form $L3$ and depending on the AT of $L3$, the last iteration is operated with either $3 \cdot \text{UNROLL}$ or UNROLL. Therefore, we inversely calculate $n$ by the multiplication of UNROLL with the number of iterations of $L3$. Thirdly, if the number of $L1$ is less than two, we need to estimate the average execution time of $L1$ by matrix multiplication. The algorithm can be found in~\cref{appendix:e}. Once we obtained all properties of $L1$, the value of $k$ is estimated in a similar fashion as above. Finally, with all estimated values, ID of DL model is estimated as shown in~\cref{alg:Algoritm1}.

\SetKwComment{tcp}{\scriptsize \textcolor{blue}{ $\triangleright$\ }}{}

\begin{algorithm2e}
 \DontPrintSemicolon

  \setcounter{AlgoLine}{0}
  \SetKwInOut{Input}{Input}
  \Input{$P$, $Q$, $UNROLL$, $stride$, $padding$, $N$(Number of iterations of loop), \\
  $ST$(Short execution time of loop), $AT$(Average execution time of loop)}
  \KwOut{$ID$: Input Dimension}

  $m \leftarrow ((N_{L2}\!-\!2)\!+\!2\!*\!(ST_{L2}/{AT_{L2}}))\!*\!P$ \tcp*[f]{{\scriptsize \textcolor{blue}{Find m}}}\

  \eIf(\tcp*[f]{{\scriptsize \textcolor{blue}{Find n}}}){$ST_{L3} < (AT_{L3}/{2})$}{
      $n \leftarrow (N_{L3}-1) * 3UNROLL + UNROLL$}{
      $n \leftarrow N_{L3}\!*\!3UNROLL$}

  \eIf(\tcp*[f]{{\scriptsize \textcolor{blue}{Find k}}}){$N_{L1} < 2$}{
  $AT_{L1} \leftarrow EstimateL1(m,n,Q)$
  
  $k \leftarrow (ST_{L1}/AT_{L1})\!*\!Q$
  }{
  $k \leftarrow ((N_{L1}-2) +2\!*\!(ST_{L1}/AT_{L1}))\!*\!Q$
  }
  $kernel \leftarrow sqrt(k/3)$
  
  $ID \leftarrow (sqrt(m)+(kernel-\!1)-\!2*padding)*stride$ 
  
  \Return $ID$
  
  \caption{Estimate Input Dimension} 
  \label{alg:Algoritm1}
\end{algorithm2e}

\subsection{SCA Results} 
\begin{figure}[t]
\begin{minipage}[b]{\columnwidth}
  \centering\raisebox{\dimexpr \topskip-\height}{%
  \includegraphics[width=0.8\textwidth]{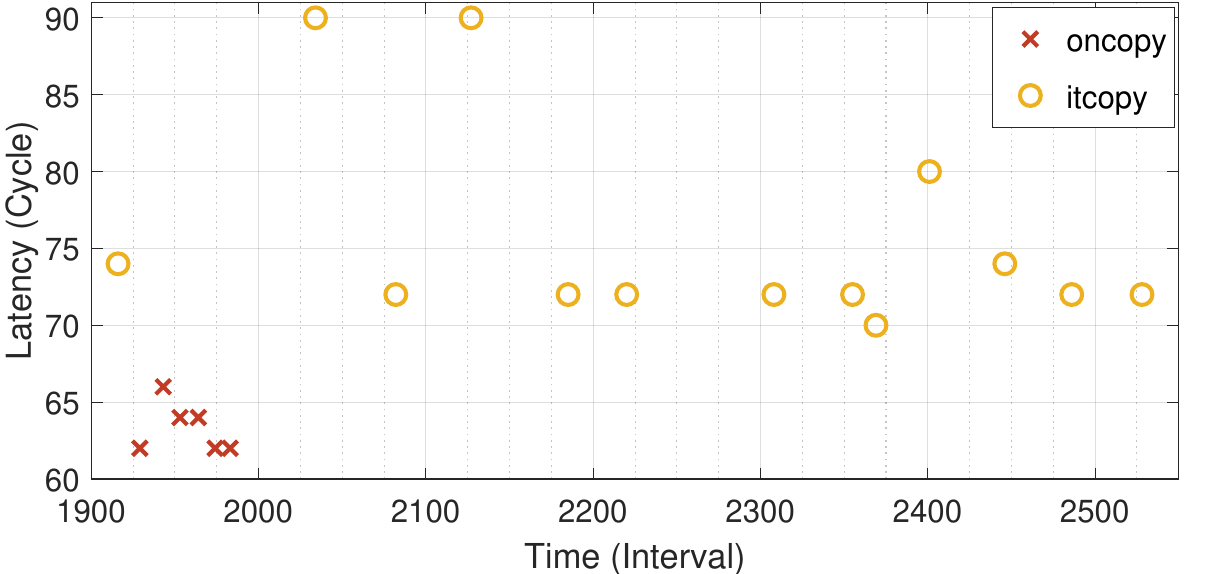}}
\caption{DCG Generation Result for \resonetwo Victim Model}
\label{fig:DCG}
\end{minipage}\hfill
\end{figure}

\textbf{Dynamic Call Graph Generation}
\revision{\cref{fig:DCG} shows the result of DCG generation after the noise filtering mechanism which only took less than 0.016 seconds to construct.} From this result, three properties described above are calculated. We note that the observations about \texttt{kernel} is omitted for brevity. Firstly, the number of each loop iteration ($N_{\text{L}}$) is calculated by counting the number of \texttt{itcopy} and \texttt{oncopy} function calls for $L2$ and $L3$ respectively. There are six \texttt{oncopy} calls between 1,900 and 2,000 time intervals and 12 \texttt{itcopy} calls after 2,000 intervals and 1 \texttt{itcopy} at the beginning.
From these, $N_{\text{L2}}$ and $N_{\text{L3}}$ are estimated as 13 and 6 respectively. $N_{\text{L1}}$ is estimated to be 1 as no calls to \texttt{oncopy} has been observed after the last \texttt{itcopy}. Also, the average and short execution time of each function call can also be estimated from the result by comparing the time intervals as described in~\cref{experimentalsetup}. All properties collected from DCG for different victim models can be found in~\cref{appendix:e}.

\begin{table}[t]
    \centering
        \caption{\revisionf{Image Dimension Estimation Result. Values in SCA and target columns represent the estimated and actual values respectively}}
    \def\arraystretch{1}
\resizebox{\columnwidth}{!}{
    \begin{tabular}{c||cc|cc|cc|cc|cc}
    
    \toprule 
    
    & 
    \multicolumn{2}{c|}{ $m$}&
    \multicolumn{2}{c|}{ $n$}&
    \multicolumn{2}{c|}{ $k$}&
    \multicolumn{2}{c|}{ $kernel$}&
    \multicolumn{2}{c}{ $ID$}
    \\ 
    \hline
    
    Victim Model&
    SCA&
    Target&
    SCA&
    Target&
    SCA&
    Target&
    SCA&
    Target&
    SCA&
    Target\\
    \hline
    \hline
    
    \resonetwo&
    4118.5&
    4096&
    72&
    64&
    35.7&
    27&
    3.5&
    3&
    129.3&
    128\\
    \bottomrule
  
    \end{tabular}%

}

    \label{tbl:sca_result}
\end{table}

\textbf{Image Dimension Estimation}
In our demonstration, pre-defined constant values of P, Q, and UNROLL are 320, 320, and 104,512 respectively. Each of $m$, $n$, and $k$ values are inversely calculated according to~\cref{alg:Algoritm1} and $kernel$ is calculated by taking the square root of $(k/3)$ (i.e., RGB channels). Finally, we round the calculated output to the nearest whole number to estimate ID of the victim model. Image dimension estimation result of different victim models are illustrated in~\cref{tbl:sca_result}. We see that even with some discrepancies in estimated values of $m$, $n$, and $k$, ID of the victim models was retrieved accurately with only ID of \resonetwo~being off only by 1.

\subsection{MEA with Model Information from SCA}
We performed subsequent MEA with estimated ID from SCA to demonstrate the benefit of using the estimated ID in MEA. We only carried out MEA only for \resonetwo~as the victim with \resonetwonine~as a surrogate model for the worst case experiment (More results in~\cref{appendix:e}). The result in~\cref{tbl:sub_mea} illustrates that \resonetwonine~(with estimated ID) achieves a slightly worse performance as ID is not the exact match. However, such performance is still better than most of MEAs with different IDs as shown in~\cref{effect_of_ID} because estimated ID was relatively closer to $I_V$. At most, the adversary can achieve up to 5.8 times better relative accuracy than the worst case \resthree. This result backs the hypothesis that employing SCA to collect model information such as ID of the model helps boost the performance of MEA.

\begin{table}[t]
    \centering
        \caption{\revisionf{Subsequent MEA (with ImageNet) result. \resonetwonine~(with estimated ID) shows slightly worse performance than \resonetwo}}
    \def\arraystretch{1.2}
\resizebox{\columnwidth}{!}{
    \begin{tabular}{c|c|c||c|c|c|c|c}

    \toprule
     \multicolumn{3}{c||}{{Victim Model}} &
     \multicolumn{5}{c}{{Surrogate Model}}\\
     \midrule
  
          \multirow{1}{*}{Dataset}&
          \multirow{1}{*}{Accuracy}&
          \multirow{1}{*}{Model}&
          \multicolumn{1}{c|}{\resthree}&
          \multicolumn{1}{c|}{\ressix}&
          \multicolumn{1}{c|}{\resonetwo}&
          \multicolumn{1}{c|}{\resonetwonine}&
          \multicolumn{1}{c}{\restwotwo}\\

    \midrule
    \hline

    \multirow{1}{*}{Indoor67} & 
    67.24\% (1x)&
    \multirow{3}{*}{\resonetwo} & 
    
    0.22x & 
    0.78x & 
    0.97x\cellcolor{gray!15} &  
    {\textbf{0.99x}}& 
    0.94x\\
    \cline{1-2}
    \cline{4-8}
   \hhline{~|~|~|~|~|-}

    \multirow{1}{*}{Caltech-256} &  
    76.75\% (1x)&
    &
    0.43x &
    0.75x &
    {\textbf{0.99x}}\cellcolor{gray!15} & 
    0.96x& 
    0.97x\\
    \cline{1-2}
    \cline{4-8}
    \hhline{~|~|~|~|~|-}
   
    \multirow{1}{*}{CUB-200} &
    77.44\% (1x)&
    &
    0.15x &
    0.59x &
    {\textbf{0.91x}}\cellcolor{gray!15} & 
    0.87x&
    0.87x \\
    
    \bottomrule
  
    \end{tabular}
}

    \label{tbl:sub_mea}
\end{table}

\section{Discussion and Limitations}
\label{discussion}

Our work targets the present and future computing environments where \revisionf{DL} models are not only running on central servers or public cloud, but also off-loaded to diverse classes of edge/endpoint devices. In these environments occur \textit{device fragmentation} referring to when users are running many different versions of software and hardware platforms. Clearly, \revisionf{DL} models running on such different platforms may also be diversified~\revision{(i.e., various \IDdot)} for efficiency or portability. \revision{Our proposed method is applicable regardless of \MA given environments where adversaries can pry on edge/endpoint devices. Moreover, with a recent development in computing power, ResNet-50 can be deployed in edge devices~\cite{qengineeringDeepLearning}.}

\textbf{Offensive Side.} Traditional MEAs armed with prior model knowledge based purely on postulations may find more challenging time to steal information from such diversified \revisionf{DL} models. Through extensive empirical analysis, we prove that a key to the success of MEA is to have the exact information about \ID and \MA of the victim model, and also demonstrated that SCAs have enough capability of extracting \ID and \MA accurately. Consequently, we suggest that for better probability of success, future researchers of MEA first should attempt to obtain the model information from SCA, instead of relying on a prior assumption.

\textbf{Defensive Side.} In our work, we quantitatively show that among model information gathered by SCA, \ID is the most essential to MEAs. This implies to defenders against MEA as well as SCA that they do not have to exhaust themselves to protect every information about their \revisionf{DL} models, but should focus mainly on concealing or obfuscating the \ID value from the adversary. For instance, to fight against SCA, they may obfuscate GEMM operations to hide actual cache access patterns. For example, dummy matrix operations may be added to the original model by inserting dummy columns and rows in the first layer of a \revisionf{DL} model to increase the number of loop iterations. Such obfuscated operations would misinform the adversary of the \ID value, which in turn eventually hampers the performance of MEA.

\textbf{Limitations.} Our study can be further improved by expanding the training datasets and architectures of victim models and utilizing newly published MEAs can revamp the analysis. In regards to SCA, due to the nature of cache-timing attack, the outcome of SCA can be obscure with noise from CPU. Such limitation may lead to repetition or failure of the attack. Therefore, devising and applying a noise filtering mechanism appropriate for the target execution environment is required in \revisionf{MEA with SCA} to maximize the performance of MEA without prior information.
\section{Conclusion}

Our systematic analysis has shown that \ID and \MA are two crucial pieces of model information as the initial knowledge about the victim for 
MEA.Our demonstration confirmed that MEA achieves the best performance when training the surrogate model with \ID identical to that of the victim model, and \MA more complex than the victim's. This result was consistent across various analysis settings. Our findings account for the reasoning behind the common design decision of existing MEA techniques that prefers their surrogates
to have the identical \IDs as the victims and as complex \MAs as possible. We note that this assumption will become unrealistic for MEAs aiming at future \revisionf{DL} models diversified to run on varied classes of computing devices because the models will have many different \IDs and \MAs depending on their devices.
However, an adversary may use advanced SCA techniques by exploiting vulnerabilities in hardware to provide fairly accurate \ID and \MA of the victim model and achieve satisfying outcomes even without such an unrealistic assumption, as demonstrated in this paper.
According to our empirical study, SCA can provide the estimated values for ID of \revisionf{DL} models that are extremely close to the target value, thereby helping the subsequent MEA to achieve the idealistic performance. From our result, defenders fighting against MEA allied with SCA may learn a lesson that they can most effectively thwart MEA by obfuscating the \ID values of their \revisionf{DL} models.

\section{Acknowledgements}
This work was supported by the BK21 FOUR program of the Education and Research Program for Future ICT Pioneers, Seoul National University in 2022
%
and the National Research Foundation of Korea (NRF) grant funded by Korea government (MSIT) (NRF-2020R1A2B5B03095204) 
\& (NRF-2022R1F1A1076100) 
and by Inter-University Semiconductor Research Center (ISRC).
Also, it was supported by Institute of Information \& communications Technology Planning \& Evaluation (IITP) grant funded by the Korea government (MSIT) 
(No.2020-0-01840, Analysis on technique of accessing and acquiring user data in smartphone) 
\& (No.2021-0-01817, Development of Next-Generation Computing Techniques for Hyper-Composable Datacenters) 
\& (No.2021-0-00724, RISC-V based Secure CPU Architecture Design for Embedded System Malware Detection and Response)

%
%
%
\appendix
\appendix
\section{Appendix}
\subsection{Model architectures}
\label{appendix:b}
\cref{apptbl:model_architecture} summarises the details of model architectures used for experiments.
\begin{table}
\centering
    \caption{Details of model architecture used for victim and surrogate models}
    \def\arraystretch{2}
\resizebox{\columnwidth}{!}{
\begin{tabular}{c||c|c|c|c|c|c|c|c|c|c|c}
\toprule

& 
\resthree&
\ressix& 
\resonetwo &
\restwotwo &
\resoneoonetwotwo&
$WRN28$-$k_{[32]}$ &
\vggthree & 
\vggsix & 
\vggonetwo &
\vggtwotwo &
\vggninetwotwo
\\ 

\hline

Parameter size&
91.65MB&
91.65MB&
91.65MB&
91.68MB&
164.13MB&
\makecell{1.43MB (k=1) 34.95MB (k=5) 139.38MB (k=10)}&
490.85MB&
516.16MB&
516.16MB&
516.16MB&
536.46MB\\



\midrule
\hline

\multirow{1}{*}{conv1} & 
($3\times3$, 64), stride 1&
($4\times4$, 64), stride 1& 
($3\times3$, 64), stride 2&
\multicolumn{2}{c|}{($7\times7$, 64), stride 2}&
($3\times3$, 16), stride 1&
\multicolumn{5}{c}{
$\begin{bmatrix}
  (\text{$3\times3$, 64})
 \end{bmatrix}$
$\times 2$
}
\\ 
\hline

\multirow{1}{*}{maxpool} & 
\multicolumn{3}{c|}{\multirow{1}{*}{\xmark}} & 
\multicolumn{2}{c|}{($3\times3$), stride 2} &
\multirow{1}{*}{\xmark} &
\multirow{1}{*}{\xmark}&
\multirow{1}{*}{\xmark}&
($2\times2$), stride 1&
\multicolumn{2}{c}{($2\times2$), stride 2}
\\
\hline 
conv2x &
\multicolumn{5}{c|}{
$\begin{bmatrix}
  (\text{$1\times1$, 64})
  (\text{$3\times3$, 64})
  (\text{$1\times1$, 256})
 \end{bmatrix}$
$\times 3$
}&
\multicolumn{1}{c|}{
$\begin{bmatrix}
  (\text{$3\times3$, $16) \times k$ }
  (\text{$3\times3$, $16) \times k$ }
 \end{bmatrix}$
$\times 4$} &
\multicolumn{5}{c}{
$\begin{bmatrix}
  (\text{$3\times3$, 128})
 \end{bmatrix}$
$\times 2$
}\\
\hline

\multirow{1}{*}{maxpool} & 
\multicolumn{5}{c|}{\multirow{1}{*}{\xmark}} & 
\multirow{1}{*}{\xmark} &
\multirow{1}{*}{\xmark} &
\multicolumn{4}{c}{($2\times2$), stride 2}
\\
\hline 

conv3x &
\multicolumn{5}{c|}{
$\begin{bmatrix}
  (\text{$1\times1$, 128})
  (\text{$3\times3$, 128})
  (\text{$1\times1$, 512})
 \end{bmatrix}$
$\times 4$
}&
\multicolumn{1}{c|}{
$\begin{bmatrix}
  (\text{$3\times3$, $32) \times k$ }
  (\text{$3\times3$, $32) \times k$ }
 \end{bmatrix}$
$\times 4$} &
\multicolumn{1}{c|}{
$\begin{bmatrix}
  (\text{$3\times3$, 128})
 \end{bmatrix}$
$\times 3$} &
\multicolumn{3}{c|}{
$\begin{bmatrix}
  (\text{$3\times3$, 256})
 \end{bmatrix}$
$\times 3$} &
\multicolumn{1}{c}{
$\begin{bmatrix}
 (\text{$3\times3$, 256})
 \end{bmatrix}$
$\times 4$}
\\
\hline 

\multirow{1}{*}{maxpool} & 
\multicolumn{5}{c|}{\multirow{1}{*}{\xmark}} & 
\multirow{1}{*}{\xmark} &
($2\times2$), stride 1&
\multicolumn{4}{c}{($2\times2$), stride 2}
\\
\hline 

conv4x &
\multicolumn{4}{c|}{
$\begin{bmatrix}
  (\text{$1\times1$, 256)}
  (\text{$3\times3$, 256)}
  (\text{$1\times1$, 1024)} 
 \end{bmatrix}$
$\times 6$
}&
{
$\begin{bmatrix}
  (\text{$1\times1$, 256)}
  (\text{$3\times3$, 256)}
  (\text{$1\times1$, 1024)} 
 \end{bmatrix}$
$\times 23$
}&
\multicolumn{1}{c|}{
$\begin{bmatrix}
  (\text{$3\times3$, $64) \times k$ }
  (\text{$3\times3$, $64) \times k$ }
 \end{bmatrix}$
$\times 4$
} &
\multicolumn{1}{c|}{
$\begin{bmatrix}
  (\text{$3\times3$, 256})
 \end{bmatrix}$
$\times 3$} &
\multicolumn{3}{c|}{
$\begin{bmatrix}
  (\text{$3\times3$, 512})
 \end{bmatrix}$
$\times 3$} &
\multicolumn{1}{c}{
$\begin{bmatrix}
  (\text{$3\times3$, 512})
 \end{bmatrix}$
$\times 4$} 
\\
\hline 

\multirow{1}{*}{maxpool} & 
\multicolumn{5}{c|}{\multirow{1}{*}{\xmark}} & 
\multirow{1}{*}{\xmark} &
($2\times2$), stride 2&
($2\times2$), stride 1&
\multicolumn{3}{c}{($2\times2$), stride 2}
\\
\hline 

conv5x &
\multicolumn{5}{c|}{
$\begin{bmatrix}
  (\text{$1\times1$, 512)}
  (\text{$3\times3$, 512)}
  (\text{$1\times1$, 2048)}
 \end{bmatrix}$
$\times 3$
}&
\multirow{1}{*}{\xmark}&
\multicolumn{4}{c|}{
$\begin{bmatrix}
  (\text{$3\times3$, 512})
 \end{bmatrix}$
$\times 3$} &
\multicolumn{1}{c}{
$\begin{bmatrix}
  (\text{$3\times3$, 512})
 \end{bmatrix}$
$\times 4$} 
\\
\hline

\multirow{1}{*}{maxpool} & 
\multicolumn{5}{c|}{\multirow{1}{*}{\xmark}} & 
\multirow{1}{*}{\xmark} &
\multicolumn{5}{c}{($2\times2$), stride 2}
\\
\hline 

&
\multicolumn{6}{c|}{averagepool, FC, Softmax}&
\multicolumn{5}{c}{maxpool, FC, FC, FC, Softmax}
\\
\bottomrule
\end{tabular}
}

\label{apptbl:model_architecture}
\end{table}

\subsection{Ablation Study}
\label{appendix:d}
\begin{figure}
\begin{subfigure}[!hb]{\columnwidth}
\centering
    \begin{subfigure}[t]{.18\columnwidth}
        \includegraphics[width=\columnwidth]{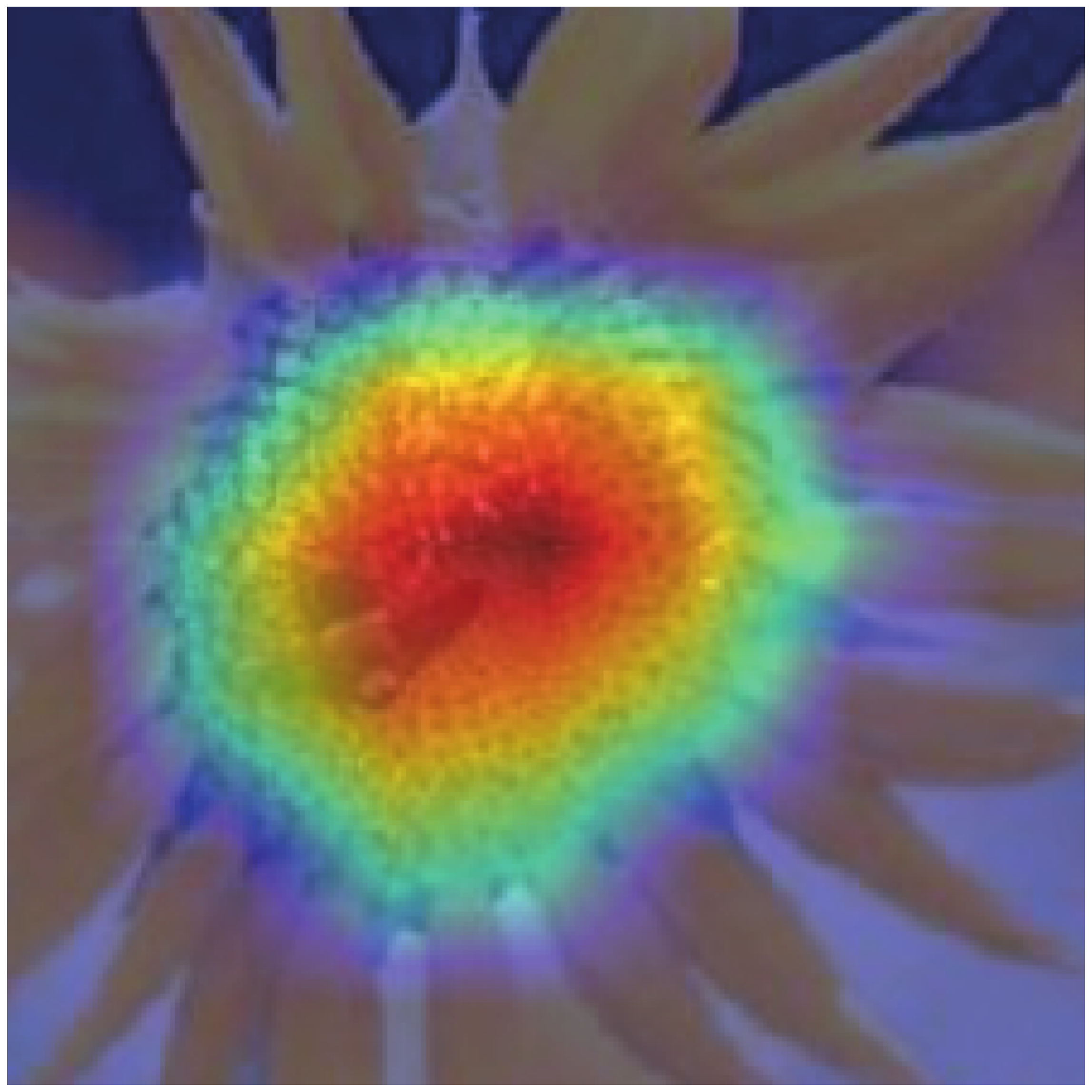}
        \captionsetup{justification=centering}
        \caption*{Victim [224] \\ Sunflower}
    \end{subfigure}
    \hspace{0.05cm}
    \vline
    \hspace{0.05cm}
    \begin{subfigure}[t]{.18\columnwidth}
        \includegraphics[width=\columnwidth]{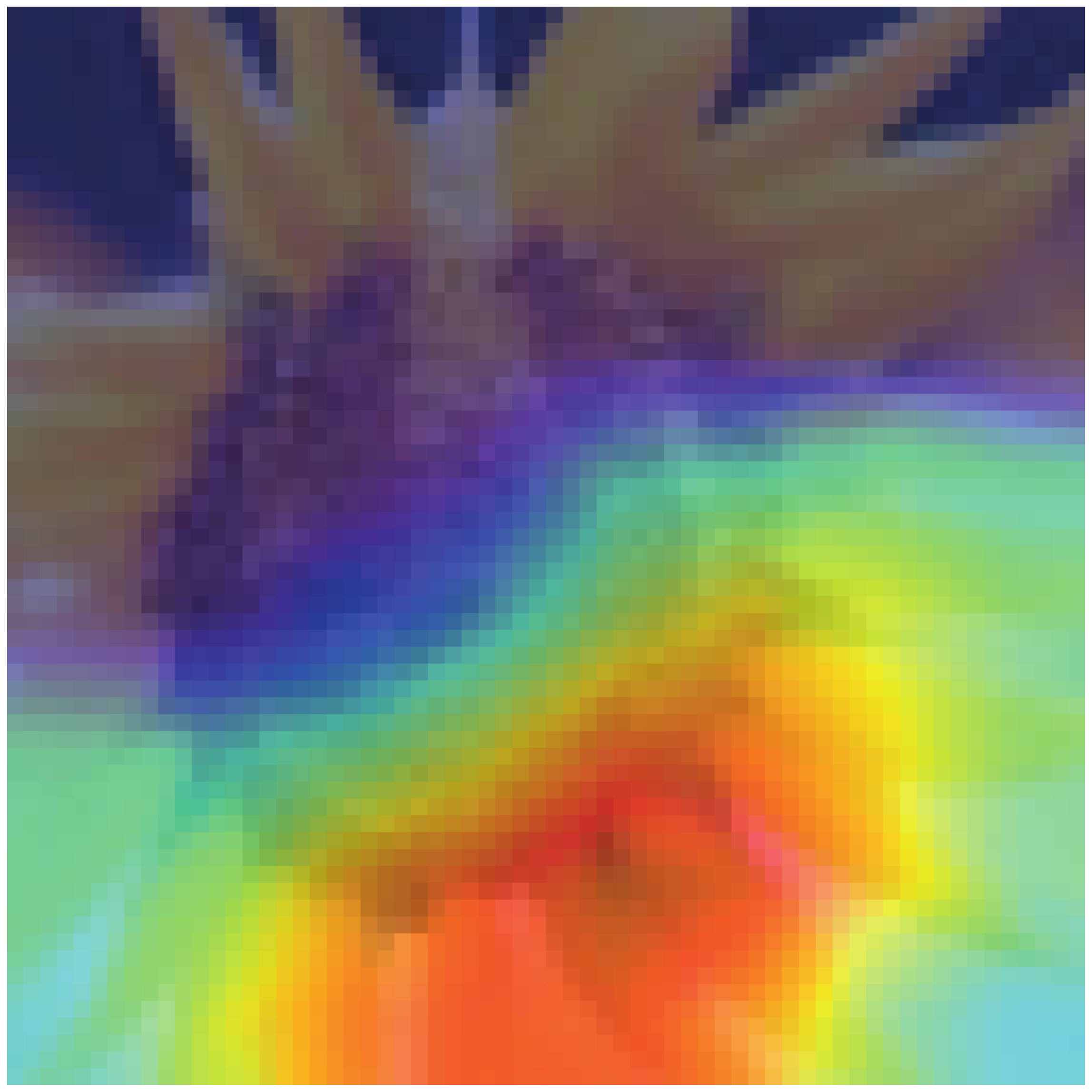}
        \captionsetup{justification=centering}
        \caption*{Surrogate [64] \\ Sunflower}
    \end{subfigure}
    \hspace{0.03cm}
    \begin{subfigure}[t]{.18\columnwidth}
        \includegraphics[width=\columnwidth]{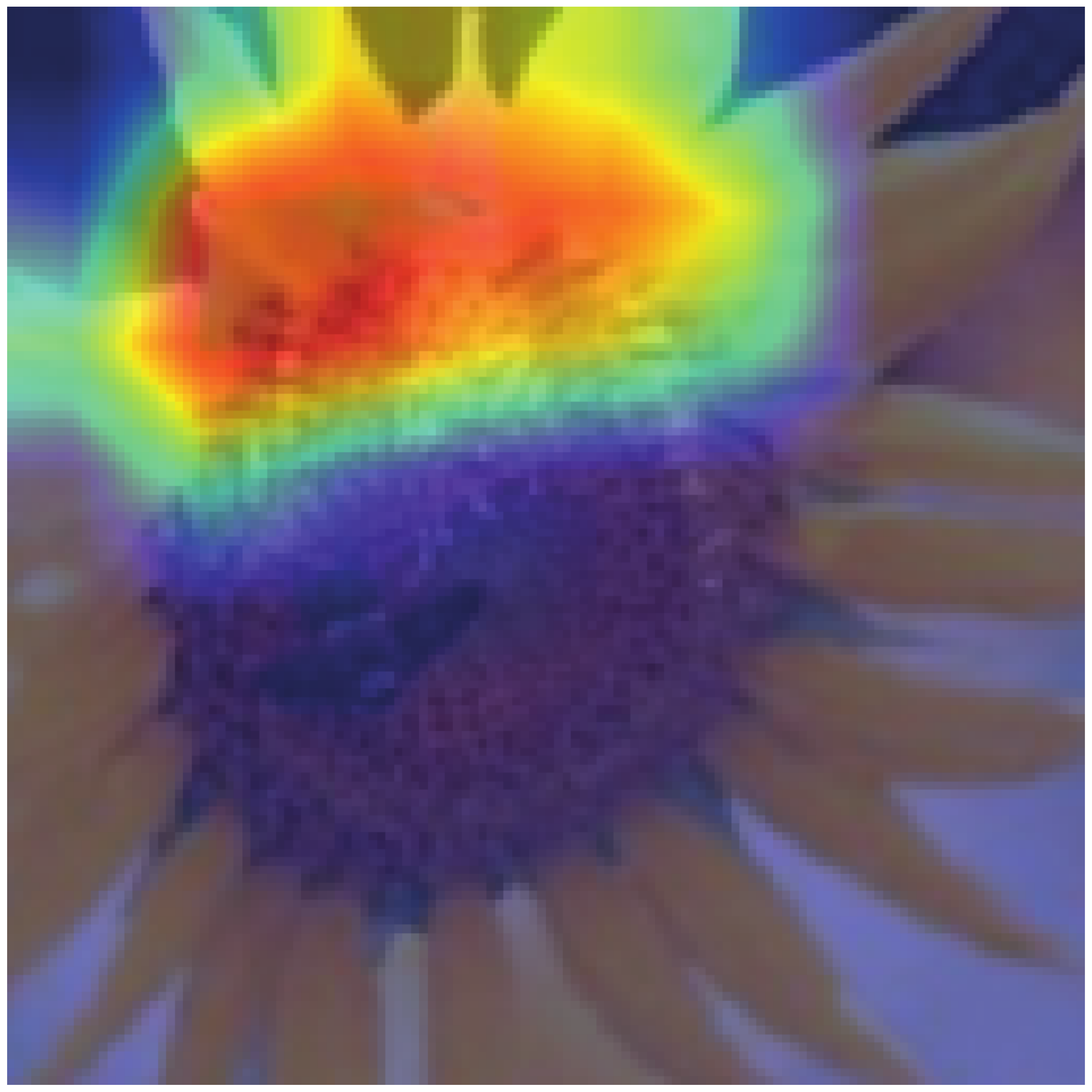}
        \captionsetup{justification=centering}
        \caption*{Surrogate [128] \\ Sunflower}
    \end{subfigure}
    \hspace{0.03cm}
    \begin{subfigure}[t]{.18\columnwidth}
        \includegraphics[width=\columnwidth]{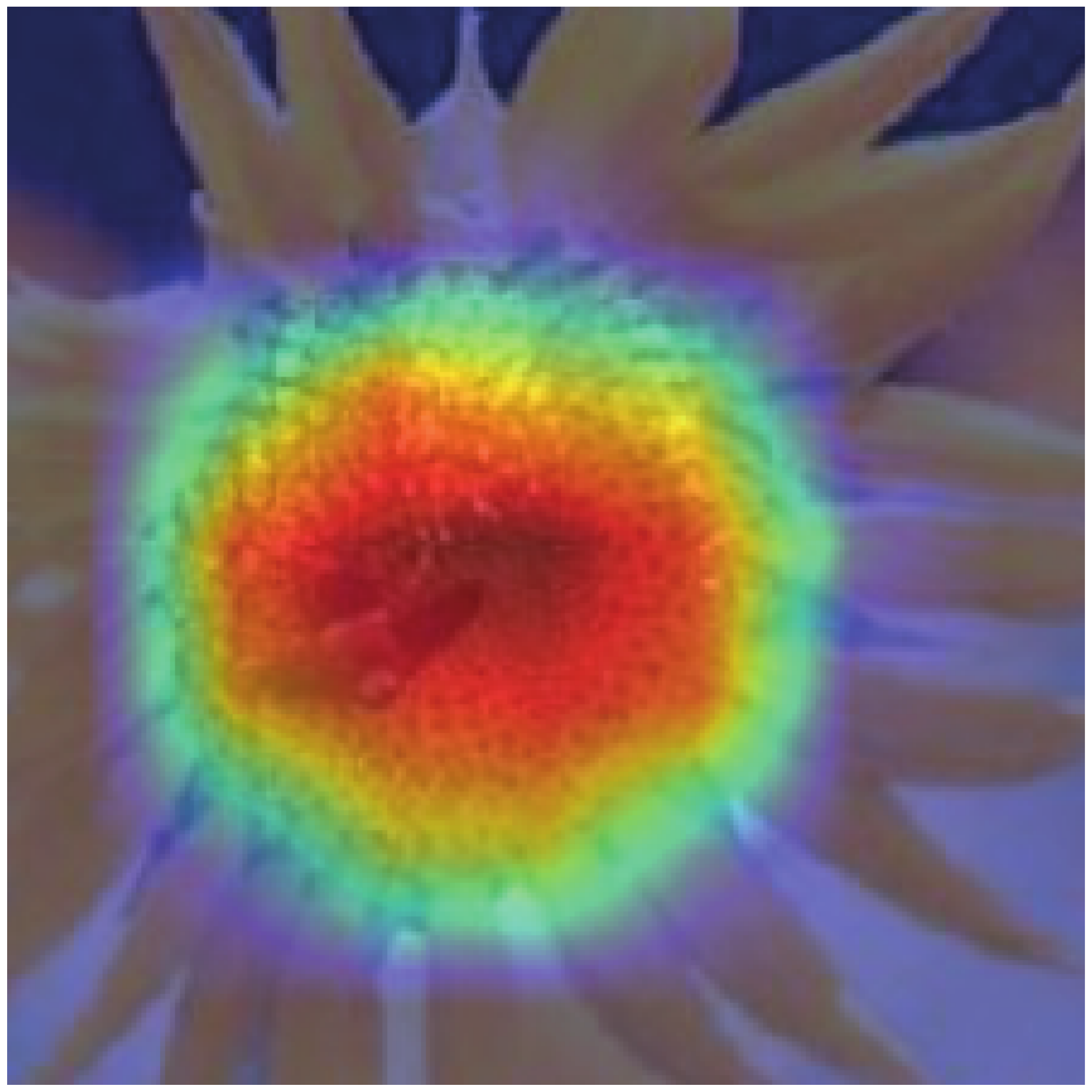}
        \captionsetup{justification=centering}
        \caption*{Surrogate [224] \\ Sunflower}
    \end{subfigure}
\end{subfigure}
\caption{Eigen-CAM Results. Only the surrogate with identical \ID is similar}
\label{appfig:grad-cam_image}
\end{figure}

\textbf{Eigen-CAM Analysis}
To further examine if the surrogate copies the victim by inheriting inner representations from convolutional layers, we carry out Eigen-CAM analysis~\cite{muhammad2020eigen} as shown in~\cref{appfig:grad-cam_image}. Even though all models predicted correctly (i.e., as sunflower), only the surrogate with the same \ID has a visual explanation similar to the victim.
\textbf{ID analysis of VGG16} An additional \ID\revisionf{analysis} with different model architecture VGG16 is shown in~\cref{apptbl:vgg_result}. ID is represented as a subscript. The result shows the same trend with the \revisionf{analysis} on \res~based model.

\begin{table}
\centering
    \caption{\revisionf{\ID \revisionf{Analysis} (Datasets) with VGG16. Effectiveness (Relative Accuracy) of MEA (KnockoffNets with ImagenNet) with query-budget-60k}}
    \def\arraystretch{1}
\resizebox{0.9\columnwidth}{!}{
    \begin{tabular}{c|c|c||cc|cc|cc|cc}
    \toprule
     \multicolumn{3}{c||}{{Victim Model}} &
     \multicolumn{8}{c}{{Surrogate Model}}\\
     \midrule
    
          \multirow{2}{*}{Dataset}&
          \multirow{2}{*}{Accuracy}&
          \multirow{2}{*}{Model}&
          \multicolumn{2}{c|}{\vggthree} &  
          \multicolumn{2}{c|}{\vggsix} & 
          \multicolumn{2}{c|}{\vggonetwo} & 
          \multicolumn{2}{c}{\vggtwotwo}  \\
          
         \cline{4-11}
          &&&
          Test 1&
          Test 2&
          Test 1&
          Test 2&
          Test 1&
          Test 2&
          Test 1&
          Test 2\\
          
    \midrule
    \hline
    
    \multirow{1}{*}{Indoor67} & 
    68.35(1x)&
    \multirow{3}{*}{\vggthree}&    
    {\textbf{0.89x}}\cellcolor{gray!15} & {\textbf{0.89x}}\cellcolor{gray!15} &
    0.54x & 0.88x&
    0.50x & 0.48x &
    0.31x & 0.20x \\

    \multirow{1}{*}{Caltech-256} & 
    73.55(1x)&
    &    
    
    {\textbf{0.94x}}\cellcolor{gray!15} & {{0.94x}}\cellcolor{gray!15} &
    0.66x & \textbf{0.96x} &
    0.62x & 0.66x &
    0.45x & 0.31x \\

     \multirow{1}{*}{CUB-200} & 
    63.82(1x)&
     
    &
    {\textbf{0.91x}}\cellcolor{gray!15} & {\textbf{0.91x}}\cellcolor{gray!15} &
    0.52x & 0.83x &
    0.49x & 0.40x &
    0.21x & 0.15x \\
    \cline{1-3}
    \cline{4-11}
    \hhline{~|~|~|~|~|-|-|}
    
    \multirow{1}{*}{Indoo67} & 
    75.00(1x)&
    \multirow{3}{*}{\vggsix}&    
    0.38x & 0.35x &
    {\textbf{0.93x}}\cellcolor{gray!15} & {\textbf{0.93x}}\cellcolor{gray!15} &
    0.73x & 0.86x &
    0.63x  & 0.63x \\
    
    \multirow{1}{*}{Caltech-256} & 
    80.55(1x)&
    &    

    0.56x & 0.51x &
    {\textbf{0.95x}}\cellcolor{gray!15} & {\textbf{0.95x}}\cellcolor{gray!15} &
    0.81x & 0.93x &
    0.78x  & 0.80x \\

     \multirow{1}{*}{CUB-200} & 
    72.56(1x)&
    &    
    
    0.24x & 0.22x &
    {\textbf{0.90x}}\cellcolor{gray!15} & {\textbf{0.90x}}\cellcolor{gray!15} &
    0.64x & 0.80x &
    0.53x  & 0.43x \\
    \cline{1-3}\cline{4-11}
    \hhline{~|~|~|~|~|~|~|-|-|}
    
    \multirow{1}{*}{Indoo67} & 
    77.91(1x)&
    \multirow{3}{*}{\vggonetwo}&    
    
    0.32x & 0.30x &
    0.70x & 0.68x &
    {\textbf{0.91x}}\cellcolor{gray!15} & {\textbf{0.91x}}\cellcolor{gray!15} &
    0.78x  & 0.81x \\
    
    \multirow{1}{*}{Caltech-256} & 
    82.39(1x)&
        
    &
    0.49x & 0.42x &
    0.81x & 0.78x &
    {\textbf{0.95x}}\cellcolor{gray!15} & {\textbf{0.95x}}\cellcolor{gray!15} &
    0.90x  & 0.92x \\
    
    \multirow{1}{*}{CUB-200} & 
    77.30(1x)&
      
    &
    0.16x & 0.13x &
    0.56x & 0.52x &
    {\textbf{0.91x}}\cellcolor{gray!15} & {\textbf{0.91x}}\cellcolor{gray!15} &
    0.74x  & 0.76x \\
     \cline{1-3}\cline{4-11}
    \hhline{~|~|~|~|~|~|~|~|~|-|-|}
    
    \multirow{1}{*}{Indoo67} & 
    78.20(1x)&
    \multirow{3}{*}{\vggtwotwo}&    
    
    0.23x & 0.27x &
    0.60x & 0.62x &
    0.84x  & 0.82x &
    {\textbf{0.92x}}\cellcolor{gray!15} & {\textbf{0.92x}}\cellcolor{gray!15} \\
    
    \multirow{1}{*}{Caltech-256} & 
    83.06(1x)&
    &    
    
    0.33x & 0.40x &
    0.76x & 0.75x &
    0.92x  & 0.91x &
    {\textbf{0.95x}}\cellcolor{gray!15} & {\textbf{0.95x}}\cellcolor{gray!15} \\
    
    \multirow{1}{*}{CUB-200} & 
    77.11(1x)&
    &    
    
    0.10x & 0.09x &
    0.38x & 0.35x &
    0.76x  & 0.71x &
    {\textbf{0.90x}}\cellcolor{gray!15} & {\textbf{0.90x}}\cellcolor{gray!15} \\

    \bottomrule
 
    \end{tabular}%

}
    \label{apptbl:vgg_result}
\end{table}

\SetKwComment{tcp}{\scriptsize \textcolor{blue}{ $\triangleright$\ }}{}
\begin{algorithm2e}
 \caption{\revisionf{$CreateDCG$}} \label{appalg:Algoritm2}
 \DontPrintSemicolon
 \KwIn{$addresses(it, on, ker), threshold$}
 \KwOut{$DCG_{a}, DCG_{d}$: Dynamic Call Graph}
 \setcounter{AlgoLine}{0}
    \For{$addr \in addresses$}{
    $delay \leftarrow probe(addr)$
    \tcp*[f]{\scriptsize \textcolor{blue}{Time taken to access addr}}
    
    ${flush(addr)}$\\
    \If(\tcp*[f]{\scriptsize \textcolor{blue}{cache hit}
    }){delay $<$ threshold}{
        ${DCG_{a}.append(addr)}$, ${DCG_{d}.append(delay)}$
        }
    }
    \KwRet $DCG_{a}, DCG_{d}$
\end{algorithm2e}

\begin{algorithm2e}
 \caption{\revisionf{$EstimateL1$}} \label{appalg:Algoritm3}
 \DontPrintSemicolon
  \KwIn{$m, n, Q, threshold$}
  \KwOut{$AT_{L1}$: $L1$ Average Execution Time}
  \setcounter{AlgoLine}{0}
  $k' \leftarrow 4Q$, $A \in \mathbb{R}^{(m,k')}$, $B \in \mathbb{R}^{(k',n)}$
  
  \lWhile{$GEMM(A, B)$}{
  $DCG_{a}, DCG_{d} \leftarrow CreateDCG()$}
  $List\ idx \leftarrow FindIndex(['itc,onc,ker,itc,ker'], DCG_{a})$
  
  $AT_{L1} \leftarrow Avg(DCG_{d}[idx][0:(idx.size()-1)])$
  
  \Return $AT_{L1}$
\end{algorithm2e}

\begin{table}
    \centering
        \caption{Properties of Loops obtained from DCG Generation Result}
    \def\arraystretch{1}
\resizebox{0.8\columnwidth}{!}{
    \begin{tabular}{c||C{1cm}C{1cm}C{1cm}|C{1cm}C{1cm}C{1cm}|C{1cm}C{1cm}C{1cm}}
    
    \toprule 
    &
    \multicolumn{3}{c|}{ $Loop1$}&
    \multicolumn{3}{c|}{ $Loop2$}&
    \multicolumn{3}{c}{ $Loop3$}\\ 
    \hline
    
    {Victim Model}&
    $N$&
    $AT$&
    $ST$&
    $N$&
    $AT$&
    $ST$&   
    $N$&
    $AT$&
    $ST$\\
    \hline
    \hline
    
    \resthree&
    1&
    1527&
    163&
    4&
    49&
    29&
    6&
    11.5&
    10\\

    \hline
    
    \ressix&
    1&
    5774&
    704&
    13&
    69.1&
    50&
    6&
    18.3&
    3.0\\
    \hline
    
    \resonetwo&
    1&
    5212&
    582&
    13&
    44.9&
    42&
    6& 	
    11.3&
    9\\
    \hline

    \restwotwo&
    1&
    17665.5&
    8163&
    40&
    208.3 &
    124&   
    6&
    38.25&
    13
    \\

    \bottomrule
  
    \end{tabular}%
}

    \label{apptbl:sca_result_loop}
\end{table}

\begin{table}
    \centering
        \caption{Image Dimension Estimation Result. Values in SCA and target columns represent the estimated and actual values respectively}
    \def\arraystretch{1}
\resizebox{0.8\columnwidth}{!}{
    \begin{tabular}{c||cc|cc|cc|cc|cc}
    
    \toprule 
    
    & 
    \multicolumn{2}{c|}{ $m$}&
    \multicolumn{2}{c|}{ $n$}&
    \multicolumn{2}{c|}{ $k$}&
    \multicolumn{2}{c|}{ $kernel$}&
    \multicolumn{2}{c}{ $ID$}
    \\ 
    \hline
    
    Victim Model&
    SCA&
    Target&
    SCA&
    Target&
    SCA&
    Target&
    SCA&
    Target&
    SCA&
    Target\\
    \hline
    \hline
    
    \resthree&
    1018.8&
    1024&
    72&
    64&
    34.2&
    27&	
    3.4&
    3&
    32.3&
    32\\
     \hline
    \ressix&
    3983.1&
    3969&
    64&
    64&
    39&
    48&
    3.6&
    4&
    63.7&
    64\\
     \hline
    \restwotwo&
    12541&
    12544&
    64&
    64&	
    147.9&
    147&
    7&
    7&
    224&
    224\\
    \bottomrule
  
    \end{tabular}%

}
    \label{tbl:sca_result_appendix}
\end{table}
\subsection{SCA algorithms and DCG generation result}
\label{appendix:e}
\cref{appalg:Algoritm2} and \cref{appalg:Algoritm3} are shown below and \revisionf{\cref{apptbl:sca_result_loop} shows the values of properties obtained from \cref{appalg:Algoritm2}. (N is the number of iterations of loop. ST and AT are the short and average execution time of loop respectively.) \cref{tbl:sca_result_appendix} shows the result of \ID estimation results for \resthree, \ressix and \restwotwo.}

\bibliographystyle{splncs04}
\bibliography{mybib}

\end{document}